\documentclass[12pt]{article}
\usepackage{amssymb,amsthm,hyperref,amsmath,graphicx,bm,color,booktabs,comment,stmaryrd}

\theoremstyle{remark}

\usepackage{float}
\usepackage{algorithm}   
\usepackage{algpseudocode}  
\usepackage{multirow}
\usepackage{subcaption}

\title{MAD-NG: Meta-Auto-Decoder Neural Galerkin Method for Solving Parametric Partial Differential Equations}
\author{ 
Qiuqi Li\thanks{E-mail: qli28@hun.edu.cn.}\\
\small{\textit{Department of  Mathematics}},\\
\small{\textit{Hunan University, Changsha, 100083, China}}
\and
Yiting Liu \thanks{E-mail: yitingliu@hnu.edu.cn.}\\
\small{\textit{Department of Mathematics}},\\
\small{\textit{Hunan University, Changsha, 100083, China}}
\and 
Jin Zhao\thanks{E-mail: zjin@cnu.edu.cn, ~Corresponding Author.}\\
\small{\textit{Academy for Multidisciplinary Studies, Beijing National Center for Applied Mathematics}},\\
\small{\textit{Capital Normal University, Beijing, 100048, China.}}
\and
Wencan Zhu \thanks{E-mail: wen@hunan.edu.cn.}\\
\small{\textit{Department of Mathematics}},\\
\small{\textit{Hunan University, Changsha, 100083, China}}
}

\date{\today}

\begin{document}
\maketitle{}

\begin{abstract}
Parametric partial differential equations (PDEs) are fundamental for modeling a wide range of physical and engineering systems influenced by uncertain or varying parameters. Traditional neural network-based solvers, such as Physics-Informed Neural Networks (PINNs) and Deep Galerkin Methods, often face challenges in generalization and long-time prediction efficiency due to their dependence on full space-time approximations. To address these issues, we propose a novel and scalable framework that significantly enhances the Neural Galerkin Method (NGM) by incorporating the Meta-Auto-Decoder (MAD) paradigm. Our approach leverages space-time decoupling to enable more stable and efficient time integration, while meta-learning-driven adaptation allows rapid generalization to unseen parameter configurations with minimal retraining. Furthermore, randomized sparse updates effectively reduce computational costs without compromising accuracy. Together, these advancements enable our method to achieve physically consistent, long-horizon predictions for complex parameterized evolution equations with significantly lower computational overhead. Numerical experiments on benchmark problems demonstrate that our methods performs comparatively well in terms of accuracy, robustness, and adaptability.
\end{abstract}

\hspace{-0.5cm}\textbf{Keywords:}
\small{Parametric partial differential equations, Deep learning, Neural Galerkin method, Meta-Auto-Decoder}\\

\section{Introduction}

In engineering, physics, biology, and other fields, parameterized dynamical systems are widely used to characterize the evolution of system behavior with parameter variations. Typical examples include magnetic field changes in electromagnetism, fluid motion in turbulent flow, and disease spread in epidemiological models.
The state of such systems is often affected by uncertain factors, such as initial conditions and system parameters. These uncertain factors can be quantified and represented using random variables. In this context, a parameterized dynamical system can be modeled as a parametric partial differential equation, where the parameters act as one or more independent variables of the system.
The ability to efficiently solve parametric PDEs has become increasingly critical as industries demand faster and more accurate solutions for real-time decision-making and optimization.

Traditional numerical methods, such as Finite Element Method (FEM), Finite Difference Method (FDM), and Finite Volume Method (FVM), are well-established. However, when applied to parameterized dynamical systems, these methods often require extensive domain discretization and repeated solving for different parameter values to accurately estimate unknown parameters and quantify their effects, leading to high computational costs. Therefore, developing efficient approaches to quantify uncertainty in parameterized dynamical systems has become an important research topic.
In the past decade, with the rapid development of deep learning techniques, many deep learning–based approaches have emerged. By incorporating physical prior knowledge into deep neural network models and leveraging the mesh-free nature of neural networks, these methods have overcome the computational bottlenecks of traditional approaches in high-dimensional spaces, offering new solutions for solving high-dimensional and complex nonlinear dynamical systems.

A series of early universal approximation theorems demonstrate that when the number of hidden neurons is sufficiently large, a neural network is capable of approximating any Borel-measurable function to an arbitrary degree of accuracy \cite{cybenko1989approximation,
hornik1989multilayer}. The past decade has witnessed a significant proliferation of neural network-based methods to solve PDEs. Prominent examples include PINNs \cite{raissi2019physics}, Deep Galerkin Methods \cite{sirignano2018dgm} and Deep Ritz Methods \cite{yu2018deep}, as well as their extensions and other influential techniques \cite{shan2025vi,zhang2020learning,han2018solving,zhang2018deep,weinan2021algorithms,zang2020weak}. These contemporary approaches build upon foundational concepts explored in pioneering early works such as \cite{dissanayake1994neural}. These approaches typically employ neural networks to approximate the full space-time solution, which can lead to a loss of temporal causality and reduced generalization performance over long time horizons. However, the solutions learned by global-in-time methods can violate causality, which can become an issue for complex problems that rely on preserving physics \cite{krishnapriyan2021characterizing}.

Due to the Dirac-Frenkel variational principle \cite{dirac1930note}, many approaches achieve the decoupling of time and space by training the network parameters solely at the initial time \cite{du2021evolutional,bruna2024neural,anderson2022evolution,berman2024neural}. 
The two works \cite{bruna2024neural,berman2024neural} introduce the NGM, which decouples space and time. It uses a neural network to approximate the solution at each time step, updates network parameters sequentially, adopts an active training data acquisition strategy, leverages the inherent properties of the current solution to achieve adaptive sampling, and constructs a dynamic loss estimation method to solve high-dimensional problems with general nonlinear parameterization. Reference \cite{berman2023randomized} proposed that the sparsity of the update can reduce the computational cost of training without sacrificing expressiveness, as many network parameters are locally redundant at each time step. In 2023, Mariella and Jans extended the paradigm of evolutionary deep neural networks (EDNNs) to the solution of parametric time-dependent partial differential equations (PDEs) on domains with geometric structures. By introducing positional embedding based on the eigenfunctions of the Laplace-Beltrami operator, they further proposed a fully training-free method, which can automatically enforce initial conditions and only requires temporal integration \cite{kast2024positional}. Additionally, Filtered Neural Galerkin model reduction schemes \cite{ning2025filtered} have been proposed for the efficient propagation of uncertainties in the initial conditions and the approach advances the mean and covariance of the solution distribution in a reduced subspace over time, achieving more than one order of magnitude speedup compared to ensemble-based uncertainty propagation.

Despite its advantages, NGM requires retraining the initial network parameters for each new parameter instance, which can be computationally expensive, particularly in the presence of stochastic or high-dimensional parameters. It is well-known that recent advances in deep learning for PDEs have shifted the focus from traditional numerical approaches to learning solution operators. These modern methods are designed to model the mapping from input functions—such as initial conditions or parameters—directly to the entire solution space. Notable examples include Deep Operator Networks (DeepONets) \cite{lu2019deeponet}, Fourier Neural Operators (FNOs) \cite{li2020fourier}, Convolutional Neural Operators (CNOs) \cite{raonic2023convolutional}, and PDE-NET \cite{long2018pde,long2019pde}, as well as other related frameworks \cite{jin2022mionet,li2024physics,hao2023gnot,bhattacharya2021model,Chen2024Positional}. Moreover, a substantial body of recent research has sought to refine these operator-learning architectures for the efficient treatment of parametrized PDEs \cite{Huang202X,Feng202X,chen2025tensor,Venturi2023SVD}. These approaches facilitate efficient generalization across diverse input configurations. In parallel, meta-learning has emerged as a potent tool within the realm of scientific machine learning \cite{psaros2022meta,finn2017model,antoniou2018train,nichol2018reptile,yoon2018bayesian,ye2024meta}. It empowers models to swiftly adapt to novel PDE instances or parameter regimes with minimal retraining. This transition from pointwise solution approximation to operator-based and meta-learned frameworks substantially enhances the applicability of deep learning in scientific computing. The MAD exemplifies this approach by leveraging meta-learning concepts to efficiently adapt to new PDE instances characterized by varying parameters, boundary conditions, and computational domain shapes \cite{ye2024meta}.

We propose the Meta-Auto-Decoder Neural Galerkin Method (MAD-NGM), an enhanced Neural Galerkin framework that incorporates the MAD paradigm. The framework is specifically designed for parameterized evolution equations with initial conditions or computational domains influenced by stochastic parameters. The key contributions of this work are:
\begin{itemize}

\item \textbf{Two-stage MAD-NGM framework for parametric PDEs.} We propose a novel two-stage MAD-NGM framework for the efficient solution of parametric PDEs, in which both the initial conditions and computational domains are influenced by stochastic uncertainties. From a methodological perspective, the proposed approach comprises two principal stages:
(1) Initial approximation stage: Within the Meta-Auto-Decoder (MAD) framework, multiple sample data are utilized to construct a nonlinear trial manifold, allowing the algorithm to rapidly infer the model parameters corresponding to the initial time.
(2) Time evolution stage: Starting from the obtained initial parameters, the Neural Galerkin Method (NGM) is employed to advance the temporal evolution of the approximate solution and to predict the system state at arbitrary time instances. Detailed parameter formulations and mathematical derivations are provided to establish the theoretical foundation of the proposed method.
\item \textbf{Randomized sparse updating strategy under the MAD-NGM framework.} When solving parametric PDEs, the MAD-NGM framework requires solving the problems and updating all network parameters at each time step, which is accompanied by temporal local overfitting and high computational costs. To address these issues, we develop the Meta-Auto-Decoder Randomized Sparse Neural Galerkin Schemes (MAD-RSNGS), which incorporate a randomized sparse update strategy that randomly updates a sparse subset of network parameters at each time step during the time evolution stage. This reduces redundant computations and significantly improves overall computational efficiency.
\item \textbf{Comparative analysis and generalization enhancement through MAD integration.} Compared with the traditional NGM, by integrating the MAD mechanism, our method eliminates the need to retrain the network from scratch for each new parameter instance. After pretraining, the model efficiently adapts to unseen parameter configurations with only a few additional training steps. Moreover, compared with the Meta-Auto-Decoder Physics-Informed Neural Network (MAD-PINN), the proposed solution framework enables physically consistent, long-term sequential predictions, demonstrating enhanced adaptability and robustness across diverse parameter regimes.

\end{itemize}

This paper is organized as follows. Section \ref{sec:Methods} introduces the problem set-up and the theoretical framework of the MAD-NGM and MAD-RSNGS. Section \ref{sec:technical implementation} presents the technical and algorithmic implementation of the proposed methods. In section \ref{sec:numerical experiments}, we present numerical results on four different examples. Section \ref{sec:conclusions} contains our concluding remarks.

\section{Mathematical Formulation of Meta-Auto-Decoder Neural Galerkin Method}
\label{sec:Methods} 
In this section, we introduce the class of parameterized partial differential equations (PDEs) that are of primary interest in this study and establish the theoretical framework and parameter representations underlying the proposed methods.
\setcounter{equation}{0}

\subsection{Evolution equations}
When solving a given partial differential equation (PDE), one typically fixes certain parameters, such as the shape of the physical domain, the diffusion or velocity field, the source term, and the flux. A PDE that depends on such parameters is referred to as a parametric PDE. When these parameters are allowed to vary over a certain range, we aim to find the solutions corresponding to all the parameter values within that range \cite{aproximate_pde}. 

Given a spatial domain $\mathcal{X} \subseteq \mathbb{R}^d$, we consider the time evolution of a field 
$u: \mathcal{X} \times [0,\infty) \rightarrow \mathbb{R}$, which, at any time $t$, belongs to a fixed function space 
$\mathcal{U}$, and satisfies the following system:
\begin{equation*} 
\left\{
\begin{aligned}
&\partial_t u(t,\boldsymbol{x}) = f(t,\boldsymbol{x},u), \quad &(t,\boldsymbol{x})\in [0,\infty) \times \mathcal{X},\\
&u(0,\boldsymbol{x}) = u_{\alpha}(\boldsymbol{x}), \quad &\boldsymbol{x} \in \mathcal{X}.
\end{aligned}
\right.
\end{equation*}

By appropriately choosing $f(t, \boldsymbol{x}, u)$, this equation can represent different equations, such as the diffusion-dominated equations. 
We assume that appropriate boundary conditions are imposed so that for any $t\in [0,\infty)$, the solution to the equation not only exists and is unique, but also depends continuously on the initial condition $u_\alpha\in\mathcal{U}(\alpha,\mathcal{X};\mathbb{R}^{d_\mathcal{U}}) = \mathcal{U}_\omega$, $\mathcal{U}_\omega\subseteq\mathcal{U}$ denotes a parameterized subspace of the function space $\mathcal{U}$ defined by the random variable $\omega = (\alpha,\mathcal{X})\in\Omega$. We assume that $\mathcal{U}_{\omega}$ is a Banach space, then we can obtain a solution map $u: \omega\to u(\omega)$ 
acting from $\Omega$ to $\mathcal{U}$. We also determine the solution manifold $\mathcal{M}=u(\Omega)=\{u(\omega),\omega\in\Omega\}$, which gathers together all solutions as the parameter varies within its range.  

\begin{figure}
    \centering
    \includegraphics[width=0.90\linewidth]{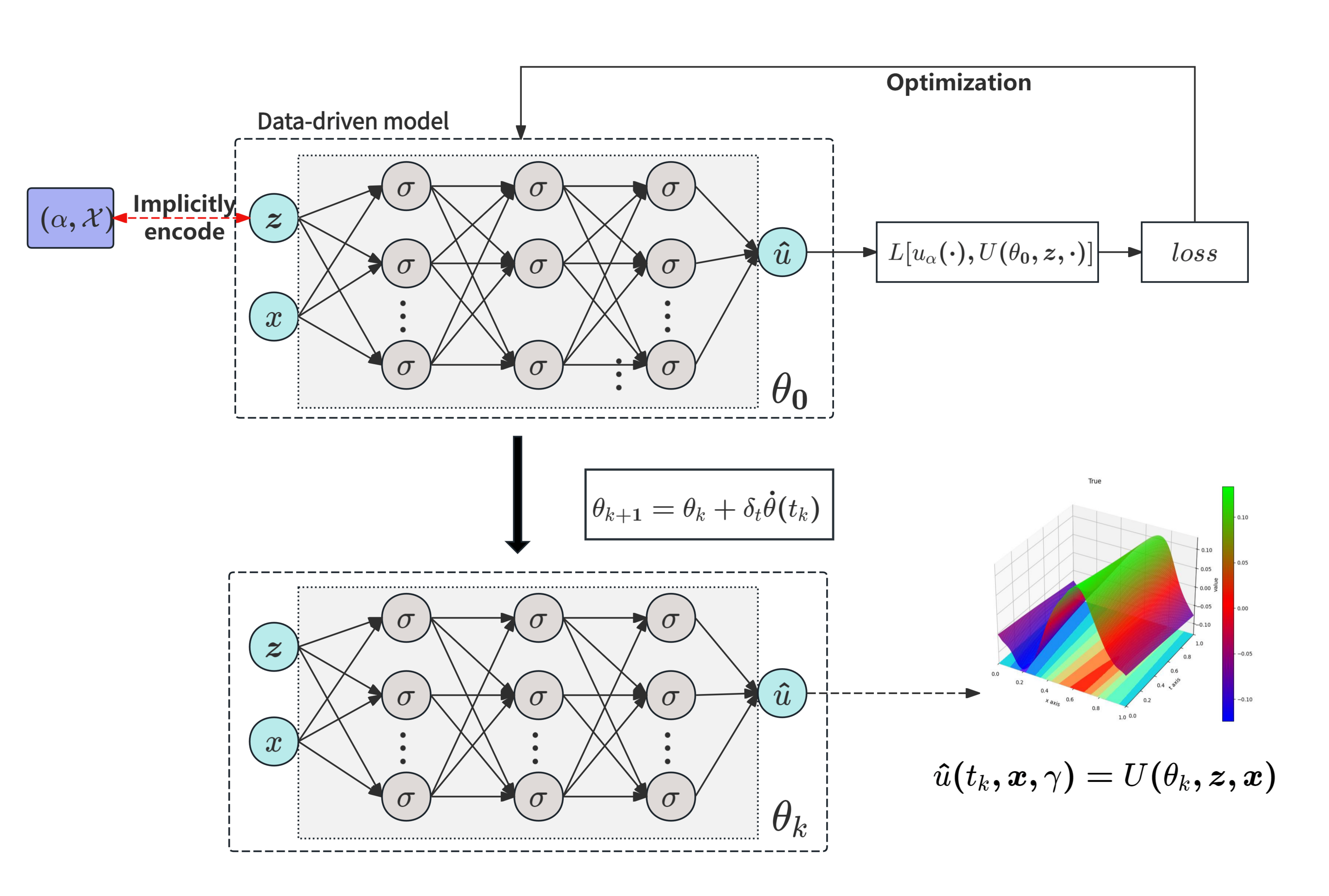}
    \caption{Architecture of the MAD-NGM framework}
    \label{fig:mad_rsngs_Method}
\end{figure}

\subsection{Approximation of Random Initial Conditions} 

To effectively approximate random initial conditions, our approach employs the MAD framework to construct a nonlinear trial manifold that represents the solution space. In this formulation, the flexibility and optimality of the trial manifold are governed by the expressive capacity of the decoder, which is primarily determined by its network width.

The set of initial conditions $\{u_{\alpha}(\boldsymbol{x})\}$ can be regarded as a compact subset $\mathcal{K} = G(\mathcal{A}) = \{ G(\omega) \mid \omega \in \mathcal{A} \} \subseteq \mathcal{U}_{\omega}$. The MAD method can construct a low-dimensional trial manifold containing $\mathcal{K}$ by training an auto-decoder model. Let $Z \subseteq \mathbb{R}^n$ be a fixed $n$-dimensional latent space. We denote by $E: \mathcal{U} \rightarrow Z$ and $D: Z \rightarrow \mathcal{U}$ the encoder and decoder mappings, respectively, both of which are assumed to be $l$-Lipschitz continuous.
For an arbitrary $n$-dimensional linear subspace $U_n \subseteq \mathcal{U}_{\omega}$, the approximation quality of $U_n$ with respect to the target set $\mathcal{K}$ can be quantified by the following metric:
\begin{equation*}
  \sup_{u \in \mathcal{K}} \|u - D(E(u))\|_{\mathcal{U}} = \sup_{\omega \in \mathcal{A}} \|u_{\omega} - D(E(u_{\omega}))\|_{\mathcal{U}} .
  \label{eqa:zhibiao1}
\end{equation*}

By taking the infimum over all admissible pairs of encoder and decoder mappings $(E, D)$, one can define a form of \emph{manifold width}, as introduced in \cite{devore1989optimal}, as follows:
\begin{equation*}
d_{n,l}(\mathcal{K}) = \inf_{\substack{E, D \\ l\text{$-Lip$}}} \ \sup_{\omega \in \mathcal{A}} \|u_{\omega} - D(E(u_{\omega}))\|_{\mathcal{U}} .
\end{equation*}
Fundamentally, the proposed approach seeks to approximate the target function set $\mathcal{K}$ by characterizing the width of the manifold and constructing a trial manifold $D(Z) = \{ D(z) \mid z \in \mathbb{R}^n \}$ through a continuous decoder mapping $D$, rather than relying explicitly on an encoder mapping $E: \mathcal{U} \to Z$. Accordingly, we define the decoder-based manifold width as
\begin{align*}
    d^{\text{Deco}}_{n,l}(\mathcal{K}) & = \inf_{\substack{D \\ l\text{$-Lip$}}} \ \sup_{u \in \mathcal{K}} \, d_{\mathcal{U}}\big(u, \{D(\boldsymbol{z}) \mid \|\boldsymbol{z}\| \le 1\} \big) \\
    &= \inf_{\substack{D \\ l\text{$-Lip$}}} \ \sup_{\omega \in \mathcal{A}} \ \inf_{\boldsymbol{z} \in Z_{B}} \|u_{\omega} - D(\boldsymbol{z})\|_{\mathcal{U}}
    \label{eq:decoder_domain_x},
\end{align*}
where $Z_{B} = \{ \boldsymbol{z} \in Z \mid \|\boldsymbol{z}\| \leq 1 \}$ denotes the closed unit ball in the latent space $Z$.
The constraint $\| \boldsymbol{z} \| \leq 1$, together with the $l-Lipschitz$ continuity imposed on the decoder mapping $D$, effectively mitigates the risk of irregular or highly oscillatory mappings, thereby ensuring enhanced stability and regularity of the resulting trial manifold.

To account for randomness in the computational domain $\mathcal{X}$, we assume the existence of a reference domain $\mathcal{X}^{\mathrm{Ref}}$ such that for any realization of $\mathcal{X}$, there exists a bijective mapping $T_{\mathcal{X}}: \mathcal{X}^{\mathrm{Ref}} \to \mathcal{X}$. For any parameter $\omega = (\alpha, \mathcal{X}) \in \mathcal{A}$, the initial condition $u_{\omega}$ defined on $\mathcal{X}$ can thus be pulled back to the reference domain as
\begin{equation*}
\hat{u}_{\omega}(\hat{\boldsymbol{x}}) 
= u_{\omega}(T_{\mathcal{X}}(\hat{\boldsymbol{x}})),\;
\hat{\boldsymbol{x}} \in \mathcal{X}^{\mathrm{Ref}},\;
\hat{u}_{\omega} \in \mathcal{U}(\mathcal{X}^{\mathrm{Ref}};\mathbb{R}^{d_{\mathcal{U}}}).
\end{equation*}
Accordingly, we define the set of transformed initial conditions as
\begin{equation*}
\mathcal{K} = \{ \hat{u}_{\omega}(\hat{\boldsymbol{x}}) \mid \omega \in \mathcal{A} \} \subset \mathcal{U}.
\end{equation*}

To accommodate possible topological variations among computational domains, we further assume the existence of a master domain $\mathcal{X}^{\mathrm{mast}}$ such that every domain $\mathcal{X}$ corresponding to $(\alpha, \mathcal{X}) \in \mathcal{A}$ satisfies $\mathcal{X} \subseteq \mathcal{X}^{\mathrm{mast}}$. Under this framework, we can define a revised notion of decoder width based on mappings defined over the master domain:
\begin{equation}
d_{n,l}^{\mathrm{Deco}}(\mathcal{K}) = \inf_{\substack{D \\ l\text{$-Lip$}}} \ \sup_{\omega \in \mathcal{A}} \ \inf_{\boldsymbol{z} \in Z_{B}} \left\| u_{\omega} - D(\boldsymbol{z})\big|_{\mathcal{X}} \right\|_{\mathcal{U}\left(\mathcal{X};\mathbb{R}^{d_{\mathcal{U}}}\right)}.
\label{equ:decoder_xxx-Tte}
\end{equation}
In this setting, we consider all $l-Lipschitz$ continuous mappings 
\[
D: Z \to \mathcal{U}(\mathcal{X}^{\mathrm{mast}}; \mathbb{R}^{d_{\mathcal{U}}}),
\]
where the decoder $D$ maps the latent space $Z$ into functions defined over the master domain $\mathcal{X}^{\mathrm{mast}}$.

\subsection{Neural Galerkin Schemes}
In this section, we introduce the Neural Galerkin Schemes \cite{bruna2024neural}. For clarity and convenience, let $u^{\alpha}(t, \boldsymbol{x})$ denote the true solution to the evolution equation corresponding to the initial condition $u_{\alpha}(\boldsymbol{x})$. We posit the existence of a time-dependent family of neural network parameters $\theta(t) \in \Theta$, which varies smoothly with time $t$. This allows the solution $u^{\alpha}(t, \cdot) \in \mathcal{U}$ at any given time $t$ to be approximated by a neural network representation.

The approximate solution, denoted by $\hat{u}^{\alpha}(t, \boldsymbol{x})$, is thus formulated as follows:
\begin{equation}
    \hat{u}^{\alpha}(t, \boldsymbol{x}) = U(\theta(t), \boldsymbol{z}, \boldsymbol{x}), \ \text{for } (t,\boldsymbol{x})\in [0,\infty) \times \mathcal{X}.
    \label{eq:nn_approximation}
\end{equation}
In this formulation, $\boldsymbol{z} \in \mathbb{R}^n$ represents a latent vector that encodes or is associated with the specific initial condition $u_{\alpha}(\boldsymbol{x})$. The function $U: \Theta \times \mathbb{R}^n \times \mathcal{X} \to \mathbb{R}$ defines the mapping of the neural network. It takes as input the time-dependent parameters $\theta(t) \in \Theta \subseteq \mathbb{R}^p$ (where $p$ is the dimensionality of the parameter space), the latent vector $\boldsymbol{z}$, and the spatial coordinate $\boldsymbol{x} \in \mathcal{X}$, and outputs the approximated solution value. The parameters $\theta(t)$ are assumed to be differentiable with respect to $t$.

Consequently, $\hat{u}^{\alpha}(t, \boldsymbol{x})$ serves as the model's approximation to the true solution $u^{\alpha}(t, \boldsymbol{x})$. Structurally, it is a nonlinear function realized by the neural network $U$, conditioned on the time-evolving parameters $\theta(t)$ and the initial-condition-specific latent vector $\boldsymbol{z}$.

In order to estimate $\dot{\theta}(t)$, we apply the chain rule to the neural network output:
$\partial U_t=\nabla_{\theta}U(\theta, \boldsymbol{z}, \boldsymbol{x})\dot{\theta}(t)$
where $\nabla_{\theta} U$ denotes the Jacobian of $U$ with respect to $\theta$.
We define the residual function at time $t$:
$$
r_t(\boldsymbol{x}, \theta(t), \boldsymbol{z}, \eta) = \nabla_{\theta(t)} U(\theta(t), \boldsymbol{z}, \boldsymbol{x}) \eta - f\big(t, \boldsymbol{x}, U(\theta(t), \boldsymbol{z}, \boldsymbol{x})\big),
$$
where $\eta \in \dot{\Theta}$ denotes a candidate value for the time derivative $\dot{\theta}(t)$, and $\dot{\Theta}$ is the set of all admissible time derivatives of $\theta(t)$.
Following the Dirac-Frenkel variational principle, the estimation of $\dot{\theta}(t)$ at time $t$ can be formulated as a least-squares problem: we seek $\eta$ that minimizes the residual in the norm, leading to the following minimization problem:
\begin{equation}
\dot{\theta}(t) \in \underset{\eta \in \dot{\Theta}}{\mathop{\arg \min}}J_{t}(\theta(t), \boldsymbol{z}, \eta),
\label{eq:minminmin}
\end{equation}
here $J_t : \Theta \times \mathbb{R}^n \times \dot{\Theta} \rightarrow \mathbb{R}$ denotes the objective function, defined in terms of the $L^2(\mathcal{X})$ norm over the spatial domain $\mathcal{X}$:
\begin{equation*}
J_{t}(\theta(t), \boldsymbol{z}, \eta)=|| r_t(\boldsymbol{x}; \theta(t), \boldsymbol{z}, \eta) ||^2_{L^2(\mathcal{X})}=||\nabla_{\theta(t)} U(\theta(t), \boldsymbol{z}, \boldsymbol{x})\eta-f(t, \boldsymbol{x};U(\theta(t), \boldsymbol{z}, \boldsymbol{x}))||^2_{L^2(\mathcal{X})}.
\label{eq:min_dou1}
\end{equation*}

The residual $\eta$ is orthogonal to the tangent space of the manifold $\mathcal{M}_{\theta} = \{U(\theta(t), \boldsymbol{z}, \cdot) \mid \theta \in \Theta\}$. If we consider the solution $\eta$ of the least-squares problem as an approximation to the desired $\dot{\theta}(t)$, we realize that the minimizers of $J_{t}(\theta(t), \boldsymbol{z}, \dot{\theta}(t)) $ satisfy the Euler-Lagrange equation:
\begin{equation*}
\nabla_{\eta}J_{t}(\theta(t), \boldsymbol{z}, \eta)=0.
\label{eq:wddwdvw}
\end{equation*}

Finally, we obtain an ODE system for the network parameters $\theta(t)$:
\begin{equation}
M(\theta(t), \boldsymbol{z})\dot{\theta(t)}=F(t,\theta(t),\boldsymbol{z}),\quad \theta(0) = \theta_0,
\label{eq:time_mmmid}
\end{equation}
here
\begin{align*}
M(\theta(t),\boldsymbol{z}) &= \int_{\mathcal{X}}\nabla_{\theta(t)}U(\theta(t),\boldsymbol{z},\boldsymbol{x})\otimes \nabla_{\theta(t)}U(\theta(t),\boldsymbol{z},\boldsymbol{x})d\boldsymbol{x},\\
F(t,\theta(t),\boldsymbol{z})&=\int_{\mathcal{X}}\nabla_{\theta(t)}U(\theta(t),\boldsymbol{z},\boldsymbol{x})f(t,\boldsymbol{x},U(\theta(t),\boldsymbol{z},\boldsymbol{x}))d\boldsymbol{x}.
\end{align*}
The initial values of the network parameters $\theta_0$ and the initial latent vector $\boldsymbol{z}$ can be determined by training based on the initial conditions.
 
\subsection{Randomized Sparse Neural Galerkin Schemes} 
In solving parameterized PDEs, the model is required not only to capture the temporal dynamics of the system but also to exhibit strong generalization and adaptability across the parameter space.
To this end, the MAD-NGM framework updates all network parameters at each time step, enabling the construction of a nonlinear trial manifold that accurately represents the temporal evolution of the solution.
This approach achieves high accuracy and strong generalization performance for a wide range of parameter configurations.

However, such a full-parameter update strategy inevitably leads to a substantial increase in computational cost.
In particular, when the dimensionality of the parameter space or the number of time steps is large, each iteration involves solving a high-dimensional least-squares problem, resulting in rapidly growing training costs and potential temporal overfitting.

To strike a balance between computational efficiency and solution accuracy, we propose an improved approach—the Meta-Auto-Decoder Randomized Sparse Neural Galerkin Schemes (MAD-RSNGS). Built upon the MAD-NGM framework, this approach incorporates a randomized sparse update strategy and establishes a dynamic system governing the evolution of neural network parameters. By introducing a sparse update mechanism for the neural network parameters $\theta(t)$, the proposed method performs random updates only on a sparse subset of parameters at each time step, thus achieving a more efficient trade-off between precision and computational complexity.

Firstly, we propose the random mapping residual. Let $e_1, \dots, e_p$ denote the standard basis vectors in $\mathbb{R}^p$, where each vector $e_i$ has the value 1 in the $i$-th entry and 0 elsewhere. Based on this notation, we define $s$ independent and identically distributed (i.i.d.) random variables $\xi_1(t), \dots, \xi_s(t)$, each depending on time $t$. Furthermore, we assume that for any $i \in \{1, \dots, s\}$, the random variables $\xi_i(t)$ are mutually independent, and each $\xi_i(t)$ takes values in the index set $\{1, \dots, p\}$ according to the distribution $\pi$.

The random mapping matrix $S_t \in \mathbb{R}^{p \times s}$ is then given by
$$
S_t = (e_{\xi_1(t)}, \dots, e_{\xi_s(t)}),
$$
where parentheses indicate column concatenation. We define the corresponding random mapping residual as
\begin{equation*}
r^s_t(\boldsymbol{x}, \theta(t), \boldsymbol{z}, \dot{\theta}_s(t)) = \nabla_{\theta(t)} U(\theta(t), \boldsymbol{z}, \boldsymbol{x}) S_t \dot{\theta}_s(t) - f\big(t, \boldsymbol{x}; U(\theta(t), \boldsymbol{z}, \boldsymbol{x})\big),
\end{equation*}
where $\dot{\theta}_s(t) \in \mathbb{R}^s$ with $s \ll p$. Here, $\nabla_{\theta(t)} U(\theta(t), \boldsymbol{z}, \boldsymbol{x})$ represents the gradient of $U(\theta(t), \boldsymbol{z}, \boldsymbol{x})$ with respect to the parameter $\theta(t)$.

Secondly, using the random mapping matrix $S_t$, we generate a sparse random projection set from the collection of component functions $\{\nabla_{\theta_i} U(\theta(t), \boldsymbol{z}, \boldsymbol{x})\}_{i=1}^p$. The matrix $S_t$ enables the extraction of a sparse subset containing only $s$ elements: 
$$
\{\partial_{\theta_{\xi_i(t)}} \hat{u}(\cdot, \theta(t), \boldsymbol{z})\}_{i=1}^s.
$$
Here, $s \ll p$, so this subset is significantly smaller than the complete set but still effectively approximates the tangent space of $\hat{u}(\cdot, \theta(t), \boldsymbol{z})$ on the parameter manifold $\mathcal{M}_{\theta}$.

Finally, to compute a suitable $\dot{\theta}_s(t) \in \mathbb{R}^s$, we formulate a least-squares problem based on the sparse generating set $\{\partial_{\theta_{\xi_i(t)}} \hat{u}(\cdot, \theta(t), \boldsymbol{z})\}_{i=1}^s$:
\begin{equation*}
\min_{\dot{\theta}_s(t) \in \mathbb{R}^s} \left\| \nabla_{\theta} \hat{u}(\cdot; \theta(t), \boldsymbol{z}) S_t \dot{\theta}_s(t) - f(\cdot; \hat{u}(\cdot, \theta(t), \boldsymbol{z})) \right\|^2_{L^2(\mathcal{X})}.
\end{equation*}

It is important to note that the choice of the distribution $\pi$ is critical, as its properties should be tailored to those of the Jacobian matrix $J(\theta(t), \boldsymbol{z})$. Since the specific distribution over the column indices can provide rigorous bounds on the sampled columns required to achieve an efficient approximation of the optimal low-rank representation of the full matrix, we adopt a uniform distribution over the index set $\{1, \dots, p\}$ in our problem. This choice is justified by the fact that the Jacobian matrices involved exhibit sufficient independence among their columns.

\section{Computational Framework and Implementation}
\label{sec:technical implementation}
\setcounter{equation}{0}
Having established the theoretical framework in the preceding sections,  we now turn to a systematic elaboration of the technical implementation of this model. Using parameter evolution laws and algorithmic mapping, a complete computational framework is constructed, enabling the effective translation of theoretical derivations into engineering practice.

\subsection{Treatment of Boundary Conditions}
In the initial condition approximation phase of the model, the primary task is to address the enforcement of the boundary condition constraints \cite{du2021evolutional,kast2024positional}. So, we employ a neural network–based positional embedding approach to enforce the boundary conditions. The principle of positional embedding is to introduce an embedding layer between the model inputs and the neural network, which performs a functional mapping of the independent variables of the equation. The mapped representations are subsequently employed as inputs to the network, ensuring that the approximate solution satisfies the prescribed boundary conditions.

For periodic boundary conditions, we suppose that the computational domain is given by $[-L, L]$.
To embed the periodic information into the neural network inputs, we adopt the following representation:
$$
[\sin\left(\frac{\pi}{L}x\right),\ \cos\left(\frac{\pi}{L}x\right),\ \boldsymbol{z}],
$$
where $x$ represents the spatial coordinate, providing explicit information about the location in space; $L$ denotes the domain size, which defines the scale of the periodic embedding and determines how the periodicity is reflected in the input.
The vector $\boldsymbol{z}$ is the latent code from the auto-decoder, which carries the underlying feature information of the data.
Through this input construction, periodic features can be clearly and explicitly expressed, ensuring that the model strictly satisfies the periodic boundary conditions during the solution process, thus improving the accuracy and reliability of the model.

\subsection{Approximation of Random Initial Conditions}
To enable supervised learning for the initial condition $u_0$, we define a loss function $\hat{L}:\mathcal{U}\rightarrow [0,\infty)$ as:
\[
\hat{L} = \|u_0(x) - U(\theta_0, \boldsymbol{z}, x)\|_{L^2(\Omega)},
\]
where $U(\theta_0, \boldsymbol{z}, x)$ denotes the prediction of the model given the initial network parameters $\theta_0$ and the latent vector $\boldsymbol{z}$.  
In practice, this loss can be approximated by:
\begin{equation}
\hat{L} \approx \frac{1}{N}\sum_{j=0}^{N-1} |u_0(x_j) - U(\theta_0, \boldsymbol{z}, x_j)|^2,
\label{eq:loss_dd}
\end{equation}
where $\{x_j\}_{j=0}^{N-1}$ is a set of sampling points in the domain $\Omega$ and $\boldsymbol{z}$ is the corresponding latent vector.

Next, we present the pretraining process. 
Let $u_{\alpha_1}(x), \dots, u_{\alpha_N}(x)$ denote the initial conditions $N$, and let the corresponding solutions be $u_{\alpha_1}(x, t), \dots, u_{\alpha_N}(x, t)$. During the pretraining phase, the model aims to extract implicit information from these $N$ samples. To this end, we refer to the previously introduced formula for decoder width (see equation \eqref{equ:decoder_xxx-Tte}), and design the following loss function:
\begin{equation}
    L(\theta, \{z_i\}_{i=1}^{N}) = \sum\limits_{i=1}^{N} \left( \hat{L}[u_{\alpha_i}(x) - U(\theta_0, \boldsymbol{z}_i, x)] + \frac{1}{\sigma} \|\boldsymbol{z}_i\|^2 \right),
    \label{eq:pretrain_loss}
\end{equation}
where $\theta$ denotes the neural network parameters, and $\boldsymbol{z}_i$ represents the latent vector associated with the $i$-th sample. Here, $\hat{L}$ quantifies the discrepancy between the predicted and true initial conditions for each sample, while the term $\frac{1}{\sigma} \|\boldsymbol{z}_i\|^2$ serves as a regularization of the latent vectors.

The pretraining process begins with the initialization of the latent vectors $N$ and the network parameters, providing an initial state for model training. Thereafter, based on the specified loss function, the latent vectors and network parameters are iteratively optimized. At each iteration, model parameters are updated in response to the loss function, enabling the model to progressively capture the meta-information inherent in the initial conditions, that is, the underlying patterns and shared features among the sample set.

The fine-tuning optimization phase follows. For a new computational task involving an unseen sample with initial condition $u_{\alpha}$, the model employs a fine-tuning optimization strategy to rapidly adapt to this sample. During this process, the neural network parameters $\theta$ remain fixed at their pretraining values, while adaptation is achieved solely by updating the model’s latent vector. Specifically, we update the latent vector by minimizing the following loss function:
\begin{equation}
L(\theta,z) = \hat{L}\left[ u_{\alpha}(x) - U(\theta, \boldsymbol{z}; x) \right] + \frac{1}{\sigma} \|\boldsymbol{z}\|^2,
\label{eq:modefine}
\end{equation}
where $\boldsymbol{z}$ denotes the latent vector associated with the new sample, and $\hat{L}$ is the loss function defined previously.

At this stage, we identify the training sample from the pretraining phase that is most similar to the current task and use its corresponding latent vector on the manifold as the initial latent vector for iteration. Meanwhile, the similarity between initial conditions is quantified through a norm-based metric to locate the index of the most similar sample. This strategy enables the model to rapidly converge to the optimal solution with only a few iterations, thereby reducing the computational cost of training for new samples.

\begin{algorithm}[H]
\caption{Training Algorithm for MAD-NGM} 
\begin{algorithmic}[1] 
\Require 
~\\
Initial conditions $\{u_{\alpha_i}(x)\}_{1\leqslant i\leqslant N}$, learning rates $\alpha_{\theta}, \alpha_{z}.$ \\
A new initial condition $u_{\alpha}(x)$, learning rate $\alpha_{z}^*.$
\Ensure 
Optimal parameters at the initial time $\theta_{new}^*$, latent vector $z_{new}^*.$

\textbf{Pre-training Stage}
\State Initialize parameters $\theta$, latent vector $\{z_{i}\}_{1\leqslant i\leqslant N};$
\For{$i = 1, 2, \ldots, N$}
    \State Compute the loss function on the basis of \eqref{eq:pretrain_loss};
    \State $\theta\leftarrow\theta-\alpha_\theta\nabla_\theta L(\theta,\{\boldsymbol{z}_i\});$
    \State $z_i\leftarrow z_i-\alpha_z\nabla_{\boldsymbol{z}_i}L(\theta,\{\boldsymbol{z}_i\});$
\EndFor

\textbf{Fine-tuning Stage}
\State $i\leftarrow\arg\min_{i}||u_{\alpha_{i}}(x)-u_{\alpha}(x)||,z\leftarrow z_{i}^{*},\theta\leftarrow\theta^{*};$
    \State Compute the loss function on the basis of \eqref{eq:modefine};
    \State $z\leftarrow z-\alpha_z^*\nabla_{\boldsymbol{z}}L(\theta,z);$

\State \Return $\theta_{new}^{*},z_{new}^{*}.$
\end{algorithmic}
\end{algorithm}

\subsection{Time Discretization}

Time discretization can be achieved using a variety of numerical schemes. In this work, we adopt the forward Euler method for illustrative purposes. Let the time domain be partitioned as $0 = t_0 < t_1 < \cdots < t_K = T$, where the time interval is $\delta t = t_{k+1} - t_k$.

\subsubsection{For the Neural Galerkin Schemes}

At each discrete time point $t_k$, let $\theta^{(k)} \in \mathbb{R}^p$ denote the discretized approximation of the continuous parameter vector $\theta(t_k)$. The initial parameter $\theta^{(0)}$, is obtained by fitting the initial condition $u_0$. The evolution of parameters in time is governed by the recursive update:
\begin{equation*}
    \theta^{(k+1)} = \theta^{(k)} + \delta t\,\dot{\theta}(t_k).
    \label{eq:tiiii}
\end{equation*}
Here, $\dot{\theta}(t_k)$ is estimated by solving the following linear system, as established in equation \eqref{eq:time_mmmid}:
\begin{equation*}
    M(\theta(t_k), z)\, \dot{\theta}(t_k) = F(t_k, \theta(t_k), z),
\end{equation*}
where $M(\theta(t_k),z)$ and $F(t_k, \theta(t_k),z)$ are typically defined by integrals involving $\nabla_{\theta} U$. In practice, these terms can be efficiently approximated using the Monte Carlo method:
\begin{align*}
    M(\theta(t_k), z) &\approx \frac{1}{n} \sum_{j=1}^n \nabla_{\theta^{(k)}} U(\theta^{(k)}, \boldsymbol{z}, x_j) \otimes \nabla_{\theta^{(k)}} U(\theta^{(k)}, \boldsymbol{z}, x_j), \\
    F(t_k, \theta(t_k),z) &\approx \frac{1}{n} \sum_{j=1}^n \nabla_{\theta^{(k)}} U(\theta^{(k)}, \boldsymbol{z}, x_j)\, f(t_k, \boldsymbol{z}, x_j, U(\theta^{(k)})),
\end{align*}
where $\{x_j\}_{j=1}^{n}$ denote the integral sampling points over the domain $\Omega$, and $\boldsymbol{z}$ represents the latent vector associated with the sample. For terms involving partial derivatives, we leverage JAX’s automatic differentiation to efficiently compute the required derivatives.

In subsequent numerical experiments, we construct the Jacobian matrix using the sampling points $\{x_j\}_{j=1}^{n}$:
\begin{align*}
J(\theta(t),z)=\begin{bmatrix}\nabla_{\theta(t)}U(\theta(t),z,x_{1}),\cdots,\nabla_{\theta(t)}U(\theta(t),z,x_{n})\end{bmatrix}^T\in\mathbb{R}^{n\times p},
\end{align*}
 which is then used to solve the following least-squares problem:
 \begin{equation}
 \min_{\dot{\theta}(t) \in \mathbb{R}^p} \left\| J(\theta(t), \boldsymbol{z}) \dot{\theta}(t) - f(\theta(t), \boldsymbol{z})) \right\|_2^2. 
 \label{least-squares-full update}
 \end{equation}

\begin{algorithm}[H]
\caption{Time Evolution Algorithm for MAD-NGM}
\label{alg:MAD-NGM} 
\begin{algorithmic}[1]
\Require 
\State Set the initial values: $\theta^{(0)} = \theta^*_{new}$, $z = z^*_{new}$.
\Ensure 
Parameters for each time level $\{\theta^{(k)}\}_{0\leqslant k\leqslant K}$, latent vector $z$.
\For{$k = 1, 2, \ldots, K$}
    \State Obtain $\{U(\theta^{(k-1)},z,x_{j})\}_{1\leqslant j\leqslant n}$ and $\{f(t,x_j,U(\theta^{(k-1)},z,x_j))\}_{1\leqslant j\leqslant n}$;
    \State Solve for sparse update $\dot\theta^{(k-1)}$ with least-squares problem\eqref{least-squares-full update};
    \State Update $\theta^{(k)} = \theta^{(k-1)} + \delta t\, \dot\theta^{(k-1)}$;
            
\EndFor 
\State \Return $\{\theta^{(k)}\}_{0\leqslant k\leqslant K}, z.$
\end{algorithmic}
\end{algorithm}

\subsubsection{For the Randomized Sparse Neural Galerkin Schemes}

For the randomized sparse variant, at each time step $t_k$, the parameter vector $\theta^{(k)}$ provides a discrete approximation to $\theta(t_k)$. The sparse selection matrix $S_k := S_{t_k}$ identifies a low-dimensional subspace in which updates are performed. The function $\hat{u}(\boldsymbol{z}, \theta(t_k), \cdot)$ serves as an approximation to $u(\cdot, t_k)$.

At each time step, the sparse update $\dot \theta_s^{(k-1)}$ serves as an estimate for the restricted time derivative $\dot{\theta}_s(t_{k-1})$, and can be calculated by solving the following least-squares problem:
\begin{equation}
    \min_{\dot\theta_s^{(k-1)}} \left\| J(\theta^{(k-1)}, \boldsymbol{z}) S_k \dot\theta_s^{(k-1)} - f(\theta^{(k-1)}, z) \right\|_2^2,
    \label{eq:min_swwwwww} 
\end{equation}
where $J_s(\theta^{(k-1)}, \boldsymbol{z}) = J(\theta^{(k-1)}, \boldsymbol{z}) S_k$ denotes the sparse Jacobian matrix at time $t_{k-1}$. The full parameter update is then recovered by mapping the sparse direction back to the original parameter space:
\begin{equation*}
    \dot \theta^{(k-1)} = S_k\, \dot\theta_s^{(k-1)}.
\end{equation*}
In practice, at each time step, the random mapping matrix is first computed, and then several columns are extracted from the Jacobian matrix to form the randomly projected Jacobian matrix. Only a selected subset of parameter directions is adapted at each step, thereby improving computational efficiency while potentially providing regularization.

\begin{algorithm}[H]
\caption{Time Evolution Algorithm for MAD-RSNGS}
\label{alg:MAD-RSNGS} 
\begin{algorithmic}[1]
\Require 
\State Set the initial values: $\theta^{(0)} = \theta^*_{new}$, $z = z^*_{new}$.
\Ensure 
Parameters for each time level $\{\theta^{(k)}\}_{0\leqslant k\leqslant K}$, latent vector $z$.
\For{$k = 1, 2, \ldots, K$}
    \State Obtain $\{U(\theta^{(k-1)},z,x_{j})\}_{1\leqslant j\leqslant n}$ and $\{f(t,x_j,U(\theta^{(k-1)},z,x_j))\}_{1\leqslant j\leqslant n}$;
    \State Get the random mapping matrix $S_k$ at the current time step;
    \State Solve for sparse update $\dot\theta_s^{(k-1)}$ with least-squares problem\eqref{eq:min_swwwwww};
    \State The full parameter update $\dot \theta^{(k-1)} = S_k\, \dot\theta_s^{(k-1)}$;
    \State Update $\theta^{(k)} = \theta^{(k-1)} + \delta t\, \dot \theta^{(k-1)}$;
            
\EndFor 
\State \Return $\{\theta^{(k)}\}_{0\leqslant k\leqslant K}, z.$
\end{algorithmic}
\end{algorithm}

\section{Numerical Experiments}
\label{sec:numerical experiments}
We demonstrate our methods on several numerical examples. We first consider the low-dimensional benchmark problems in Section \ref{kdv} and Section \ref{burgers} to demonstrate the learning and generalization capabilities of the models. Subsequently, we perform a comparative analysis between our methods and the meta-learning-based PINN approach. In Section \ref{allen-equation}, we progressively extend our method to problems where both the solution domain and initial conditions involve uncertainties. Finally, we further extend the framework to high-dimensional spatial domains. In the entire experimental section, the mean squared error (MSE) is adopted as the evaluation metric for error calculation at specific time $t_j$:
\begin{equation*}\label{eq:error}
    MSE=\frac{1}{N_{test}}\frac{1}{N}\sum_{i=1}^{N_{test}}\sum_{i=1}^{N}(u_i^j-u_i^{*,j})^2,
\end{equation*}
where $N_{test}$ is the size of the testing set, $N$ is the number of spatial grid points.

\subsection{Korteweg-de Vries Equation:}
\label{kdv}
\setcounter{equation}{0}
We consider a Korteweg-de Vries (KdV) equation \cite{miles1981korteweg} defined in the spatial domain $\Omega = [-1, 1]$ and in the temporal interval $T = [0, 1]$:
\begin{equation*}\label{eq:pde}
    u_t + u u_x + \frac{1}{400} u_{xxx} = 0,
\end{equation*}
The solution is subject to the following initial and boundary conditions:
\begin{itemize}
    \item[\textbf{(BC)}] \textbf{Periodic boundary condition:} $u(-1,t) = u(1,t),\quad t \in [0,1]$;
    \item[\textbf{(IC)}] \textbf{Initial condition:} $u(x,0) = \alpha_1 \sin(\pi x) + \alpha_2 \cos(\pi x)$, where $\alpha_1, \alpha_2 \in [-0.5,\, 0.5]$.
\end{itemize}

To make the model naturally satisfy the periodic boundary conditions, the boundary embedding layer we choose
$[\sin(\pi x),\cos(\pi x),\boldsymbol{z}].$ 
In this experiment, 200 initial condition samples are generated.
We adopt a single-layer neural network architecture comprising 20 neurons with the tanh activation function, and set the dimension of the latent vector to 5. We choose to adopt the pseudospectral method from the Chebfun library to generate the reference solution, which serves as a reliable accuracy benchmark for the model's calculation results.
During pretraining, the model is trained on the training set for 1,000 iterations with the L-BFGS optimizer to jointly optimize the network parameters and latent representations, thus establishing the initial model weights as well as the experimental manifold context.  
Temporal evolution is simulated using the classical fourth-order Runge–Kutta method \cite{butcher1996history} over 1,000 time steps, up to the final time $t = 1$.

\subsubsection{Results of the Training Set Samples}

\begin{figure}[H]
\centering 
    \begin{minipage}[c]{0.49\textwidth}
    \centering
    \includegraphics[width=\textwidth]{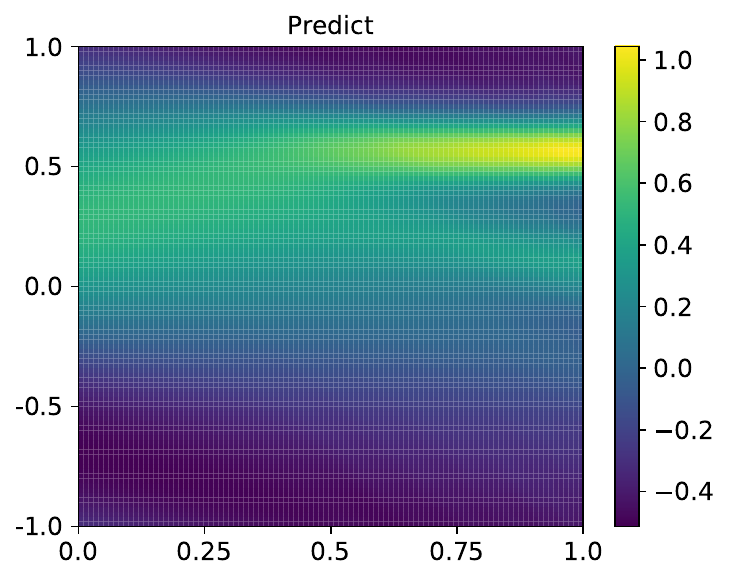}
    \label{Fig.subfig_2_1}
    \end{minipage}
    \centering
    \begin{minipage}[c]{0.49\textwidth}
    \centering
    \includegraphics[width=\textwidth]{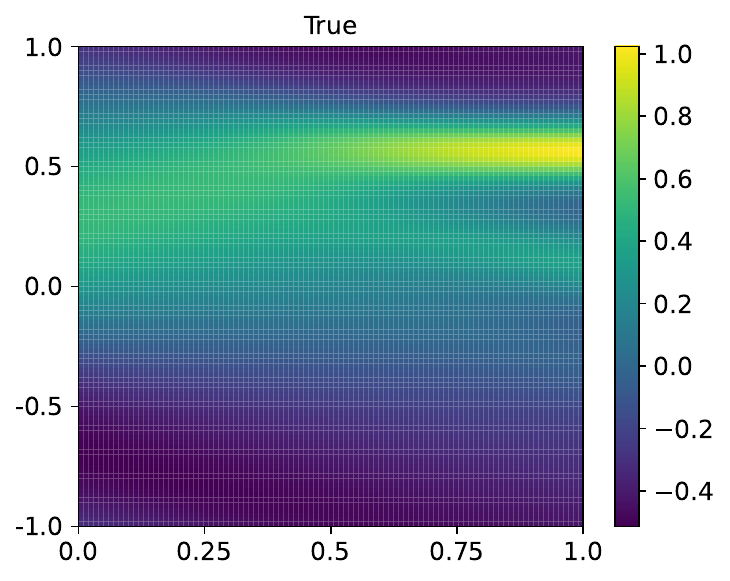}
    \label{subfig_2_2}
    \end{minipage}\\
\caption{Comparison of predicted and reference solutions for the KdV equation on the training set, showing that the proposed approach effectively captures both the nonlinear and dispersive characteristics of the wave evolution.} 
\label{Fig2.kdv_0_res_tttt} 
\end{figure}

\begin{figure}[H]
\centering 
    \begin{minipage}[c]{0.32\textwidth}
    \centering
    \includegraphics[width=\textwidth]{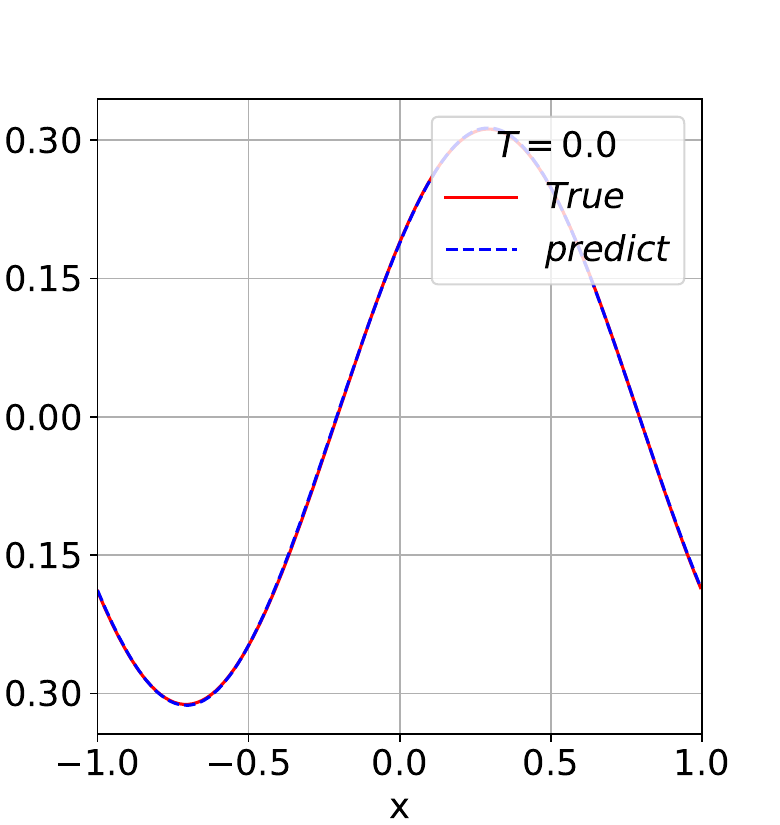}
    \end{minipage}
    \centering
    \begin{minipage}[c]{0.32\textwidth}
    \centering
    \includegraphics[width=\textwidth]{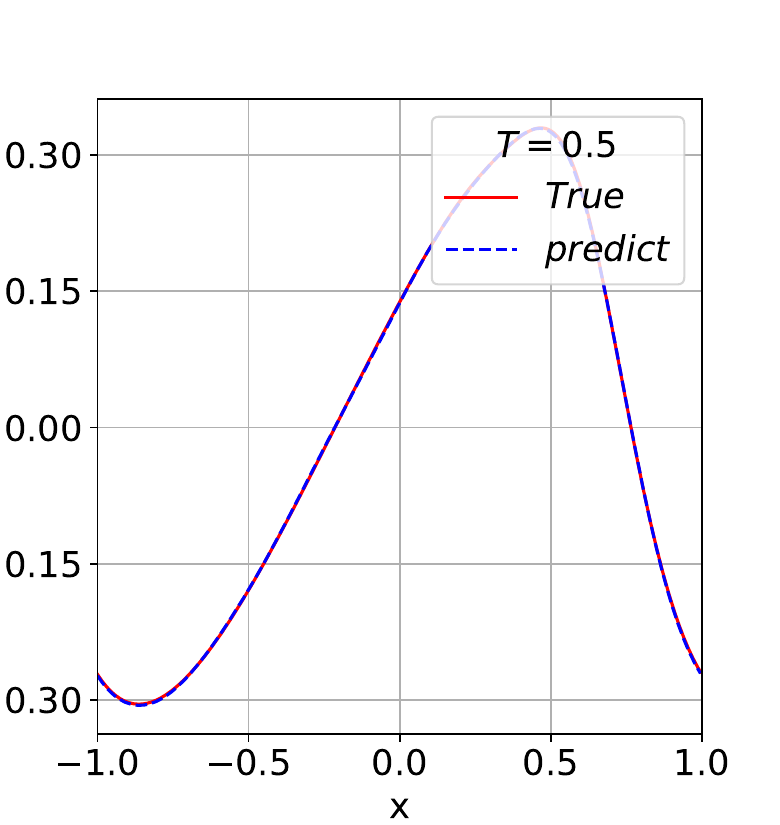}
    \end{minipage}
    \centering
    \begin{minipage}[c]{0.32\textwidth}
    \centering
    \includegraphics[width=\textwidth]{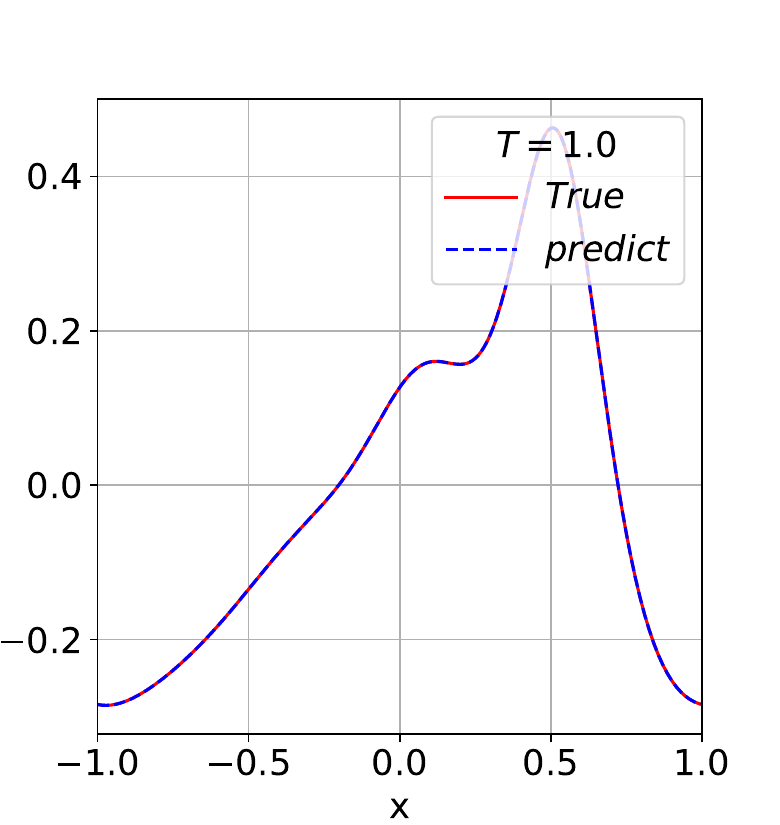}
    \end{minipage}\\
\caption{Temporal evolution of the reference and predicted solutions for the KdV equation on the training sample at $t = 0.0~,0.5~,1.0$, illustrating the high reconstruction accuracy of the MAD-NGM method for nonlinear dispersive wave dynamics.} 
\label{Fig3.kdv_0_res_tttt} 
\end{figure}

Here, a sample is selected from the training set, and 400 fine-tuning iterations are performed based on the proposed algorithm to obtain the initial latent vector. During the subsequent time evolution, a full update approach is used to update the network parameters. The related results can be seen in Figure~\ref{Fig2.kdv_0_res_tttt} and \ref{Fig3.kdv_0_res_tttt}.

\subsubsection{Results of the Test Set Samples}

\begin{figure}[H]
\centering 
    \begin{minipage}[c]{0.49\textwidth}
    \centering
    \includegraphics[width=\textwidth]{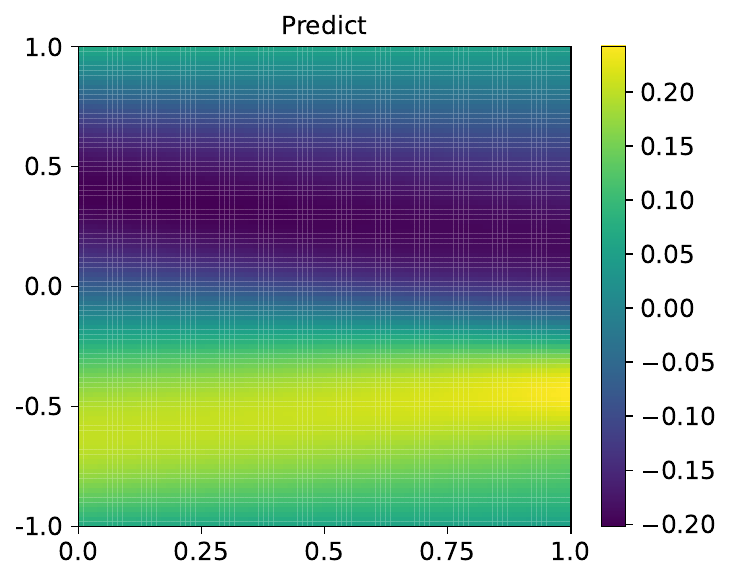}
    \label{Fig.subfig_4_1}
    \end{minipage}
    \centering
    \begin{minipage}[c]{0.49\textwidth}
    \centering
    \includegraphics[width=\textwidth]{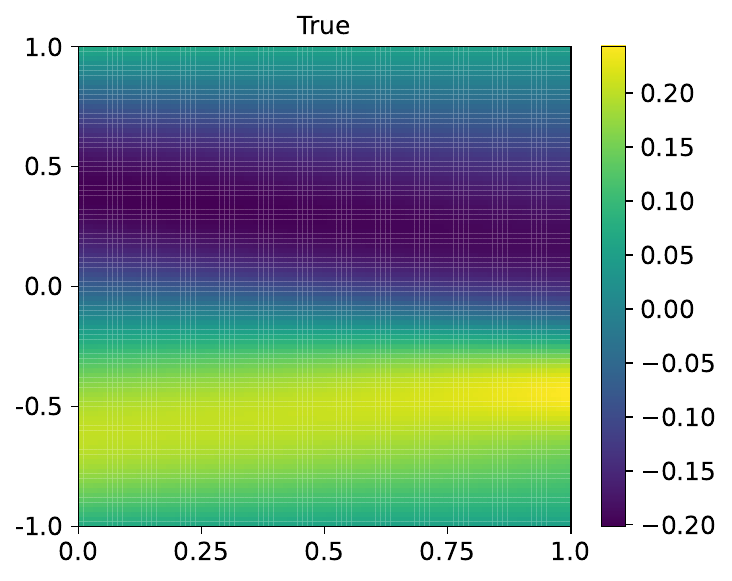}
    \label{subfig_4_2}
    \end{minipage}\\
\caption{Comparison of predicted and reference solutions for the KdV equation on the testing set, demonstrating the strong generalization and predictive capability of the proposed MAD-NGM framework.} 
\label{Fig4.kdv_198_res_t} 
\end{figure}

\begin{figure}[H]
\centering 
    \begin{minipage}[c]{0.32\textwidth}
    \centering
    \includegraphics[width=\textwidth]{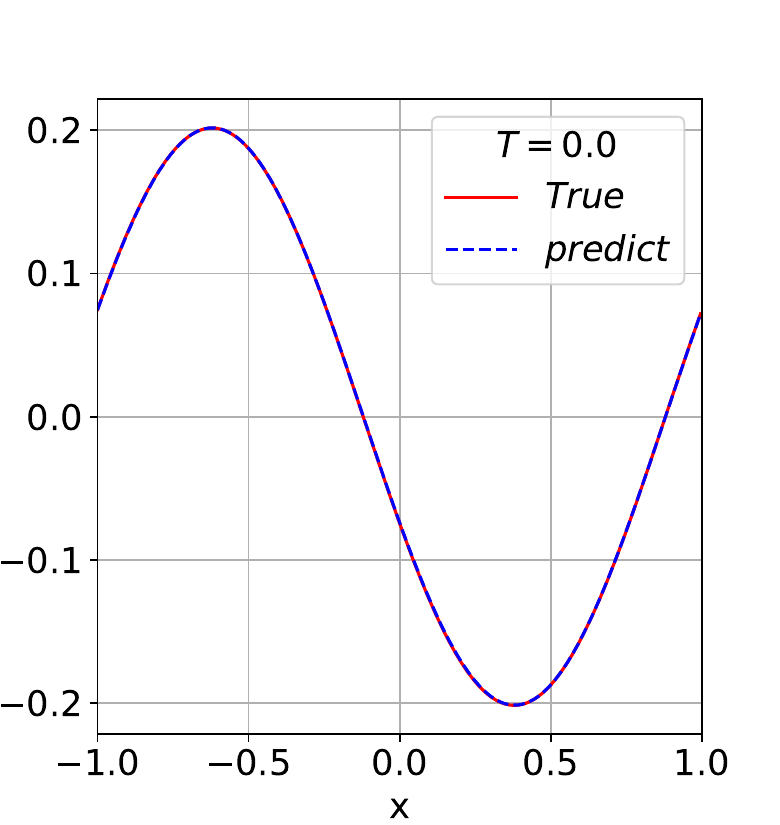}
    \label{Fig.subfig_5_1}
    \end{minipage}
    \centering
    \begin{minipage}[c]{0.32\textwidth}
    \centering
    \includegraphics[width=\textwidth]{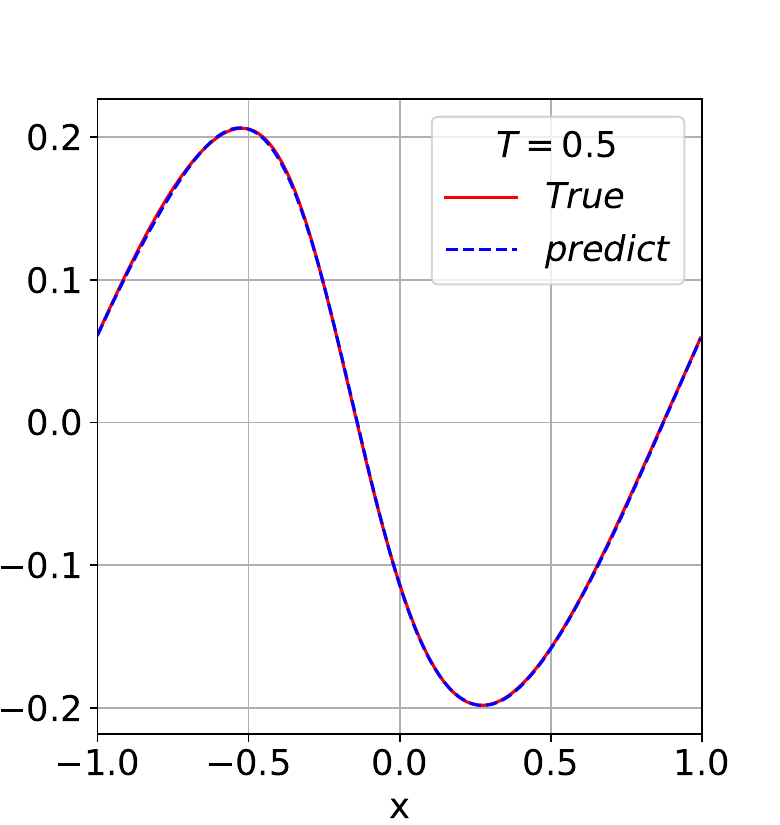}
    \label{Fig.subfig_5_2}
    \end{minipage}
    \centering
    \begin{minipage}[c]{0.32\textwidth}
    \centering
    \includegraphics[width=\textwidth]{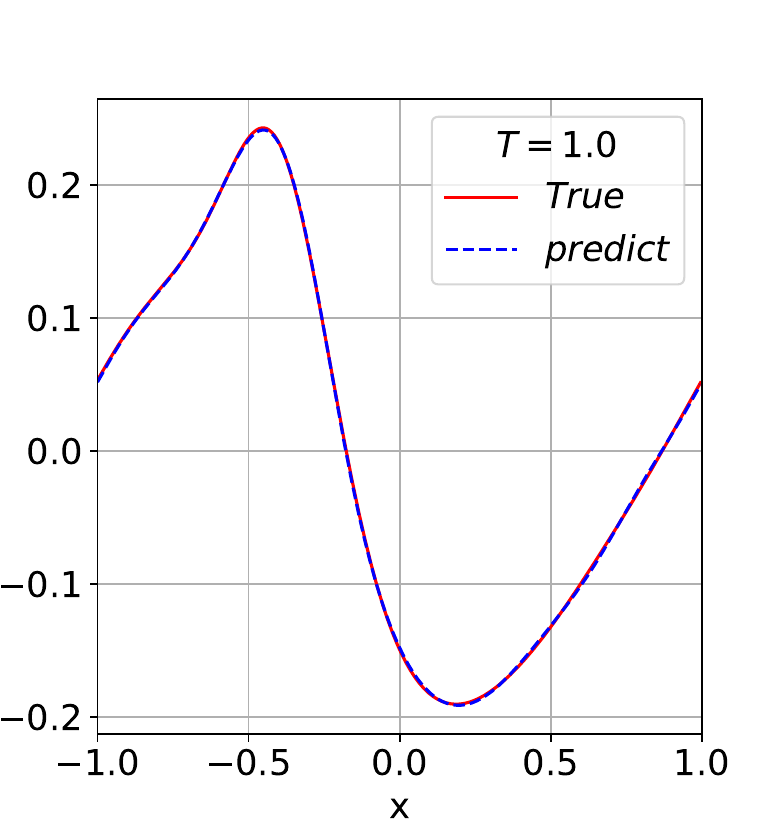}
    \label{subfig_5_3}
    \end{minipage}\\
\caption{Comparison of predicted and reference one-dimensional profiles of the KdV equation at $t = 0.0~,0.5~,1.0$ on the testing sample. The results confirm the reliability of the proposed MAD-NGM framework in reproducing the testing dynamics.} 
\label{Fig5.kdv_198_res_tttt} 
\end{figure}

For a new computational task, fine-tuning is performed based on the closest pre-trained network parameters. The model is trained for 400 iterations, followed by time evolution to obtain the approximate solution. The corresponding results can be seen in Figure~\ref{Fig4.kdv_198_res_t} and \ref{Fig5.kdv_198_res_tttt}.

To ensure fair comparison, both experimental groups use identical pre-trained training and testing samples. Each model is fine-tuned for 100, 400, and 4000 iterations, respectively. The approximation error measured by mean squared error (MSE) is assessed at the start and end of the fine-tuning and at selected time points ($t = 0.2$, $t = 0.5$, and $t = 1.0$) during temporal evolution.

\begin{table}[h] 
    \centering 
    \caption{MSE obtained at different fine-tuning iterations and time instances for both the training and test datasets. The fine-tuning stage significantly improves accuracy, and the error remains within the same order of magnitude for unseen samples, confirming the model’s generalization performance.} 
    \label{tab:kdv_ex2_1}
    \setlength{\tabcolsep}{2pt} 
    \begin{tabular}{ccccccc}
    \toprule
    \textbf{Dataset} & \textbf{Iterations} & \textbf{Pre-trained} & \textbf{Fine-tuned} & \boldmath{$t=0.2$} & \boldmath{$t=0.5$} & \boldmath{$t=1.0$} \\
    \midrule
    \multirow{3}*{Training set} 
    & 100  & $6.46\times 10^{-7}$ & $6.28\times 10^{-7}$ & $1.47 \times 10^{-6}$ & $1.05 \times 10^{-5}$ & $1.25\times 10^{-4}$ \\
    ~ & 600  & $6.46\times 10^{-7}$ & $6.24\times 10^{-7}$ & $1.47 \times 10^{-6}$ & $1.04 \times 10^{-5}$ & $6.27\times 10^{-5}$ \\
    ~ & 4000 & $6.46\times 10^{-7}$ & $5.74\times 10^{-7}$ & $1.56 \times 10^{-6}$ & $1.19 \times 10^{-5}$ & $3.69\times 10^{-5}$ \\
    \midrule
    \multirow{3}*{Test set}  
    & 100  & $4.95\times 10^{-4}$ & $2.71\times 10^{-7}$ & $2.84 \times 10^{-7}$ & $6.88 \times 10^{-7}$ & $1.04\times 10^{-5}$ \\
    ~ & 600  & $4.95\times 10^{-4}$ & $2.63\times 10^{-7}$ & $2.75 \times 10^{-7}$ & $6.71 \times 10^{-6}$ & $1.02\times 10^{-5}$ \\
    ~ & 4000 & $4.95\times 10^{-4}$ & $1.17\times 10^{-6}$ & $1.22 \times 10^{-6}$ & $1.72 \times 10^{-6}$ & $9.02\times 10^{-6}$ \\
    \bottomrule
    \end{tabular}
\end{table}

Based on the preceding experiments and Table~\ref{tab:kdv_ex2_1}, the following conclusions can be drawn:
(1) The model exhibits rapid convergence on new samples, demonstrating strong generalization capability;
(2) The accuracy of the numerical solution during temporal updates is closely related to the accuracy at the initial time step. In general, higher initial accuracy leads to improved performance in subsequent steps, and this correlation holds within a certain range;
(3) The meta-learning approach is capable of capturing the implicit distribution of samples in the manifold. During fine-tuning, only the latent vector needs to be adjusted, enabling the model to achieve satisfactory performance in both fine-tuning and subsequent temporal updates.

\subsection{Burgers Equation:} 
\label{burgers} 
We consider a Burgers equation \cite{bonkile2018systematic} defined over the spatial domain $\Omega = [0, 1] $ and the temporal interval $T = [0, 1] $:
\begin{equation*}
\frac{\partial u}{\partial t}+u\frac{\partial u}{\partial x} = \frac{1}{100\pi}\frac{\partial^2u}{\partial x^2},
\end{equation*}
The solution is subject to the following initial and boundary condition:
\begin{itemize}
\item[\textbf{(BC)}] Periodic boundary condition: $ u(0,t) = u(1,t), \ t \in [0,1] $;
\item[\textbf{(IC)}] Initial condition: $ u(x,0) \sim \mathcal{N}\left(0, 7^2 \left(-\Delta + 7^2 I\right)^{-3}\right) $, where the random field satisfies the periodic boundary condition.
\end{itemize}


In this experiment, the model employs the MAD-RSNGS framework, with pre-training performed on 100 samples. The latent vector dimension is set to 80 and the neural network consists of four layers, each comprising 30 neurons. During the pre-training phase, the L-BFGS optimizer is used to optimize the initial network parameters and latent vectors over 17,000 iterations.
In the fine-tuning phase, for each specific sample, the Adam optimizer with a learning rate of 0.01 is applied for 1,800 iterations to determine its corresponding initial latent vector.
During the sparse update process, the time step is set to 0.001, and the simulation proceeds until the final time $t = 1.0 $. At each time step, 450 network parameters are randomly selected for update. 
                 
\begin{figure}[H] 
\centering 
    \begin{minipage}[c]{0.49\textwidth}
    \centering
    \includegraphics[width=\textwidth]{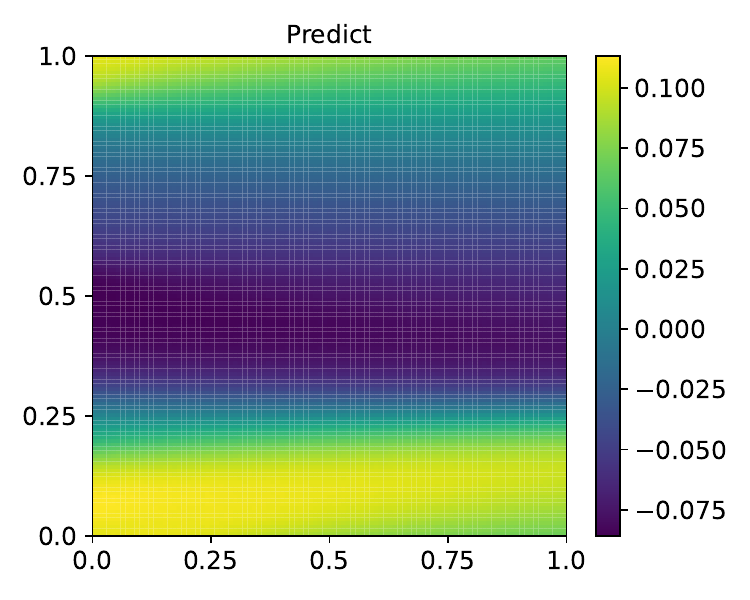}
    \label{Fig.subfig_6_1}
    \end{minipage}
    \centering
    \begin{minipage}[c]{0.49\textwidth}
    \centering
    \includegraphics[width=\textwidth]{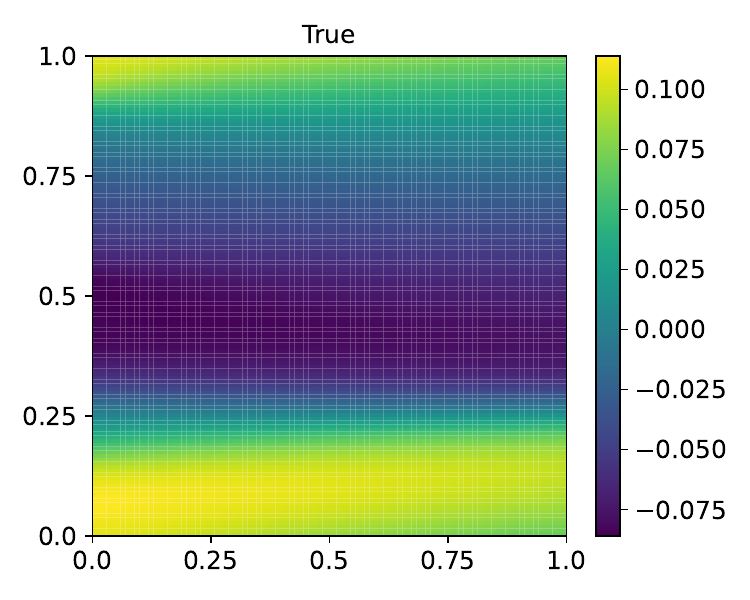}
    \label{subfig_6_2}
    \end{minipage}\\
\caption{Comparison between the predicted and reference solutions of the Burgers equation obtained by the proposed MAD-RSNGS. The prediction accurately captures the amplitude and spatial features without artifacts. } 
\label{Fig.burgers_001_pi_res_ng_time} 
\end{figure} 

\begin{figure}[H]
\centering 
    \begin{minipage}[c]{0.32\textwidth}
    \centering
    \includegraphics[width=\textwidth]{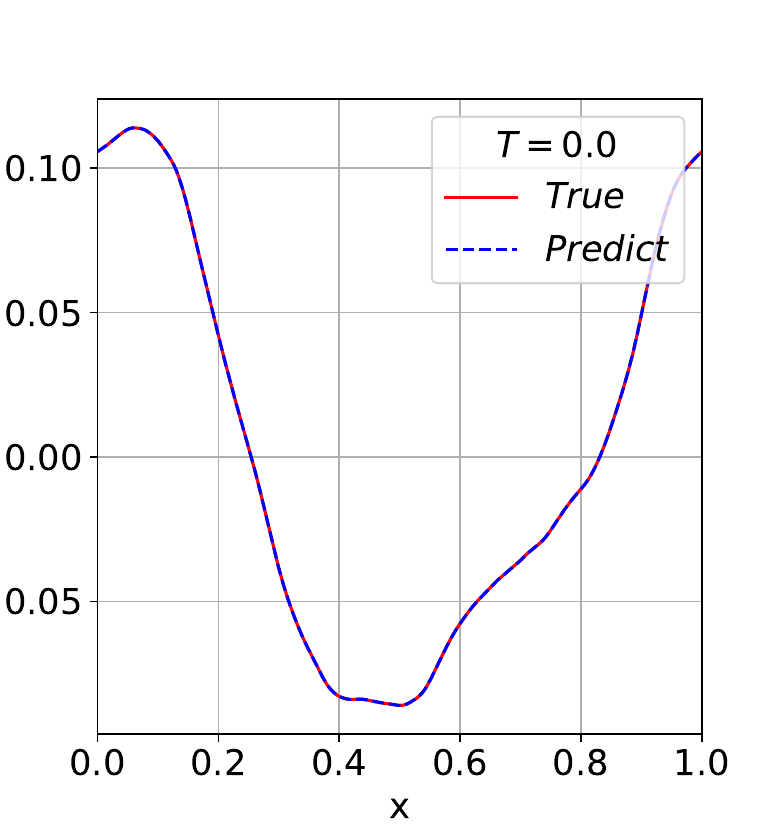}
    \label{Fig.subfig_7_1}
    \end{minipage}
    \centering
    \begin{minipage}[c]{0.32\textwidth}
    \centering
    \includegraphics[width=\textwidth]{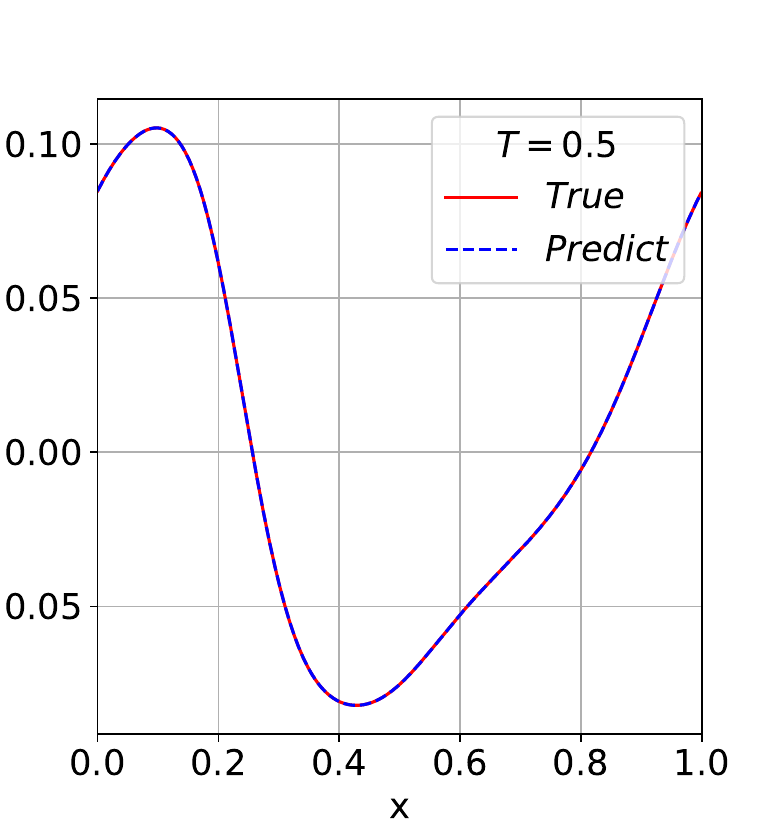}
    \label{Fig.subfig_7_2}
    \end{minipage}
    \centering
    \begin{minipage}[c]{0.32\textwidth}
    \centering
    \includegraphics[width=\textwidth]{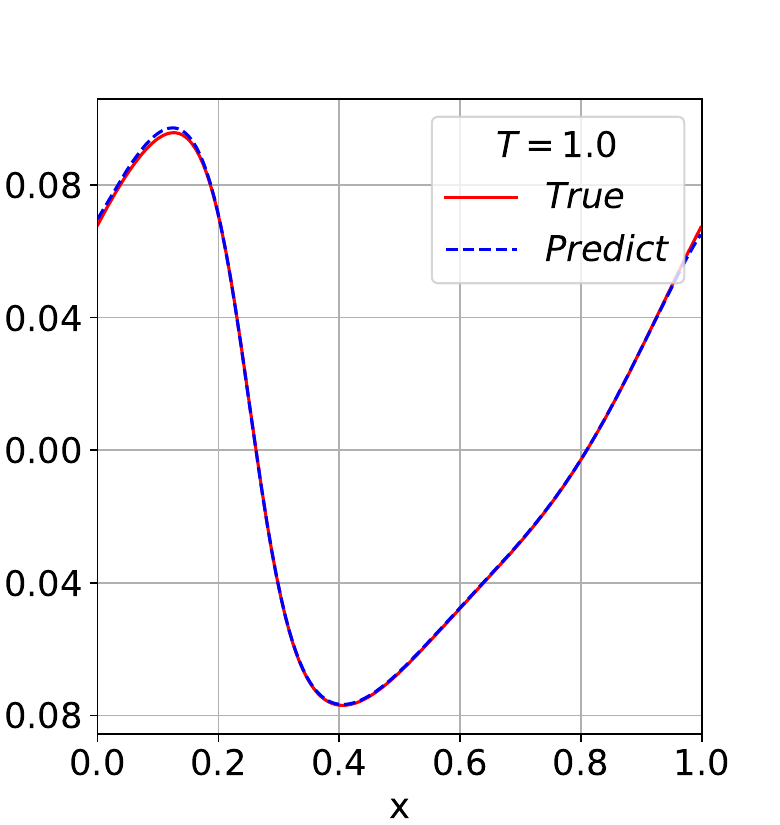}
    \label{subfig_1_2}
    \end{minipage}\\
\caption{Numerical solutions of Burgers equation obtained through MAD-RSNGS. Plots show spatial profiles at $t = 0.0~,0.5~,1.0$. The randomized sparse update strategy preserves solution accuracy while reducing computational overhead. } 
\label{Fig.kdv_198_res_tttt} 
\end{figure}

To evaluate the accuracy of the proposed method, we compare the approximate solution with the reference solution at different time points:
$t = 0.0,~ 0.5,~ 1.0$.
As depicted in Figure~\ref{Fig.burgers_001_pi_res_ng_time},~\ref{Fig.kdv_198_res_tttt}, the solutions generated by MAD-RSNGS exhibit excellent agreement with the reference solutions in the spatial domain $[0,1]$. Specifically, leveraging a randomized sparse update strategy, the proposed method not only accurately captures the shock formation and propagation dynamics but also successfully resolves the nonlinear wave steepening and shock structure inherent to the Burgers equation, thereby yielding high-precision approximations.

\subsubsection{Comparative Evaluation of Different Methods} 

This section highlights the differences among the MAD-NGM, MAD-RSNGS, and MAD-PINN. The experiments are conducted on 100 samples generated under the same parameter settings as in the previous example, from which 80 samples are selected to form the training dataset. For MAD-RSNGS and MAD-PINN, they also follow the same network architecture and optimization procedures as MAD-NGM mentioned above.

In the inference phase of MAD-RSNGS, 514 sample points are selected to approximate the initial condition. The temporal integration is carried out with an effective time-step size corresponding to 1,100 optimization iterations, advancing the solution to $t = 1.1$, At each integration step, a subset of 450 network parameters is randomly selected. 
In contrast, the MAD-PINN constructs its loss function using 13,029 sample points distributed throughout the entire space-time solution domain.

\begin{figure}[http]
    \centering 
    
    \begin{minipage}[c]{0.45\textwidth} 
        \centering
        \includegraphics[width=\textwidth]{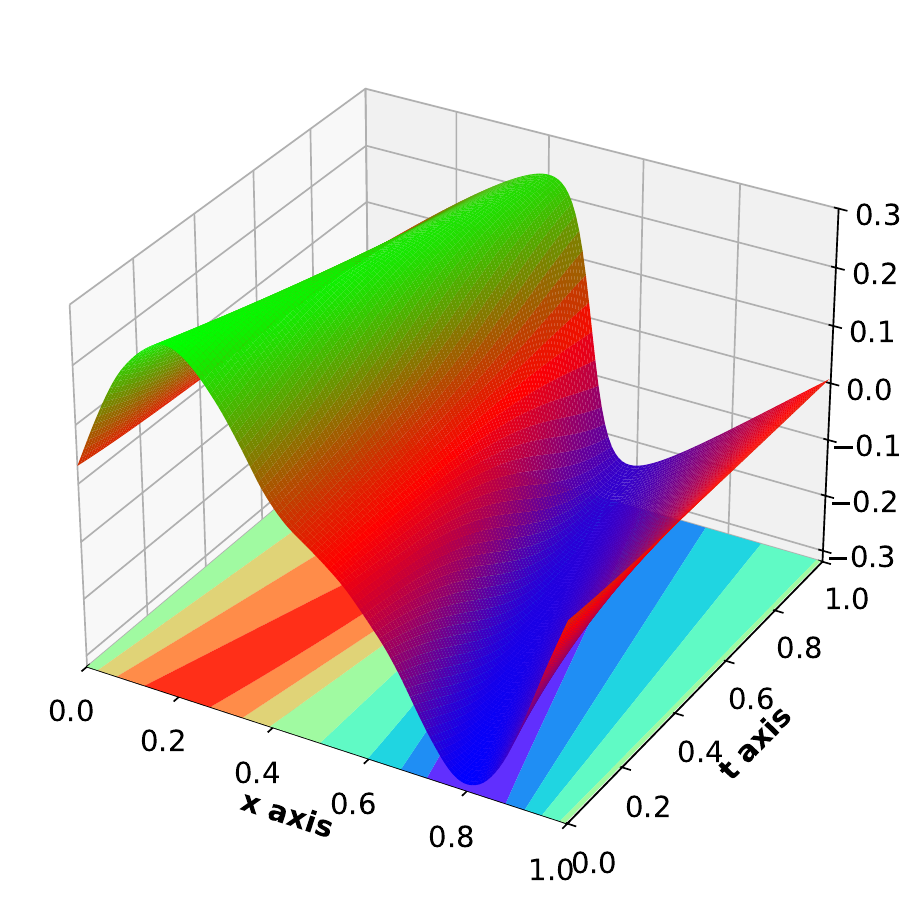} 
        \subcaption{True}
        \label{Fig.subfig_8_1}
    \end{minipage} 
    \hfill 
    \begin{minipage}[c]{0.45\textwidth}
        \centering
        \includegraphics[width=\textwidth]{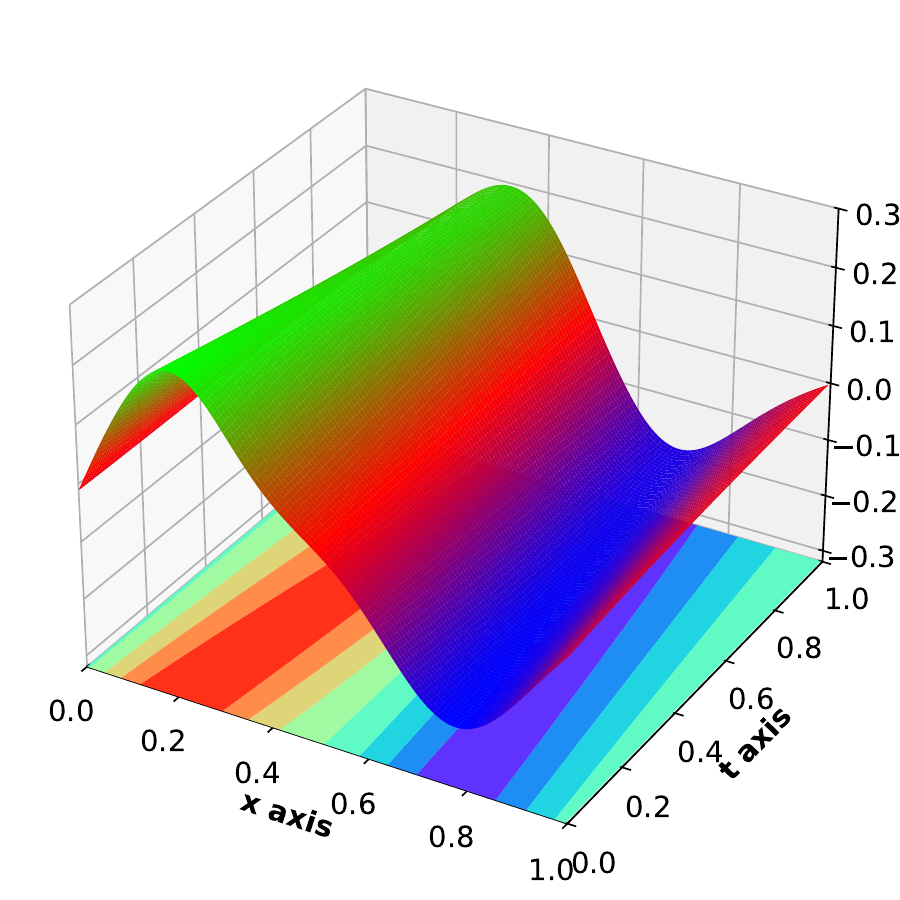}
        \subcaption{MAD-PINN}
        \label{Fig.subfig_8_2}
    \end{minipage}

    \begin{minipage}[c]{0.45\textwidth}
        \centering
        \includegraphics[width=\textwidth]{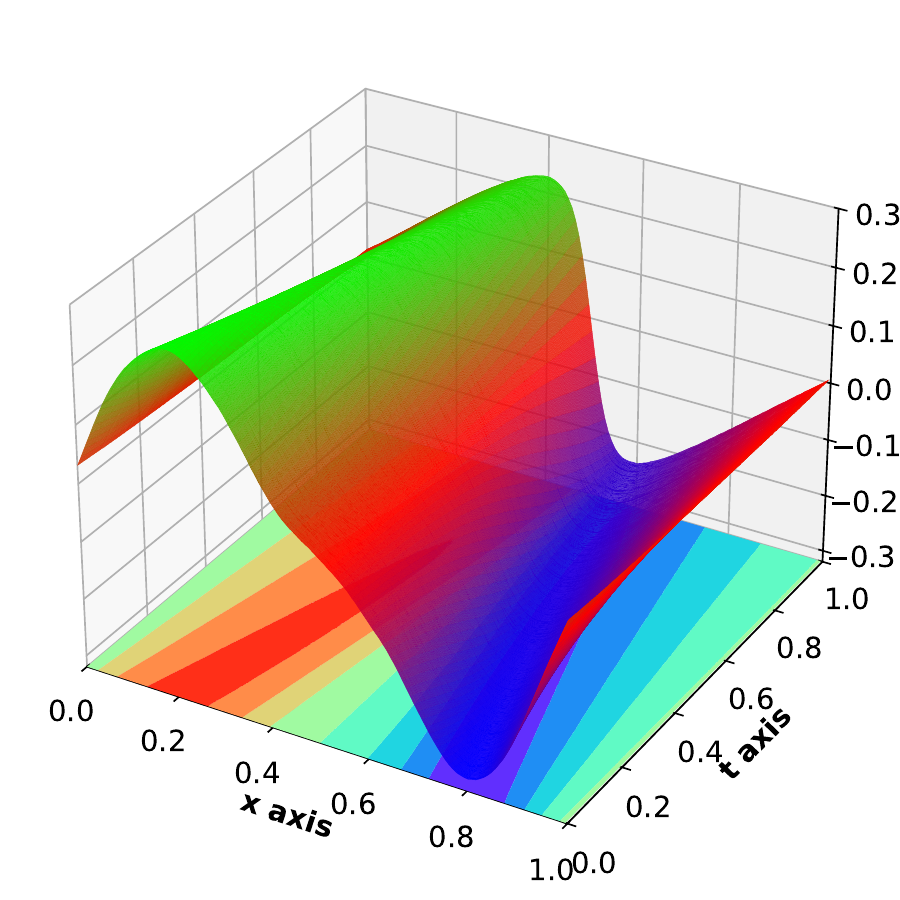} 
        \subcaption{MAD-NGM}
        \label{Fig.subfig_8_3} 
    \end{minipage} 
    \hfill 
    \begin{minipage}[c]{0.45\textwidth} 
        \centering
        \includegraphics[width=\textwidth] {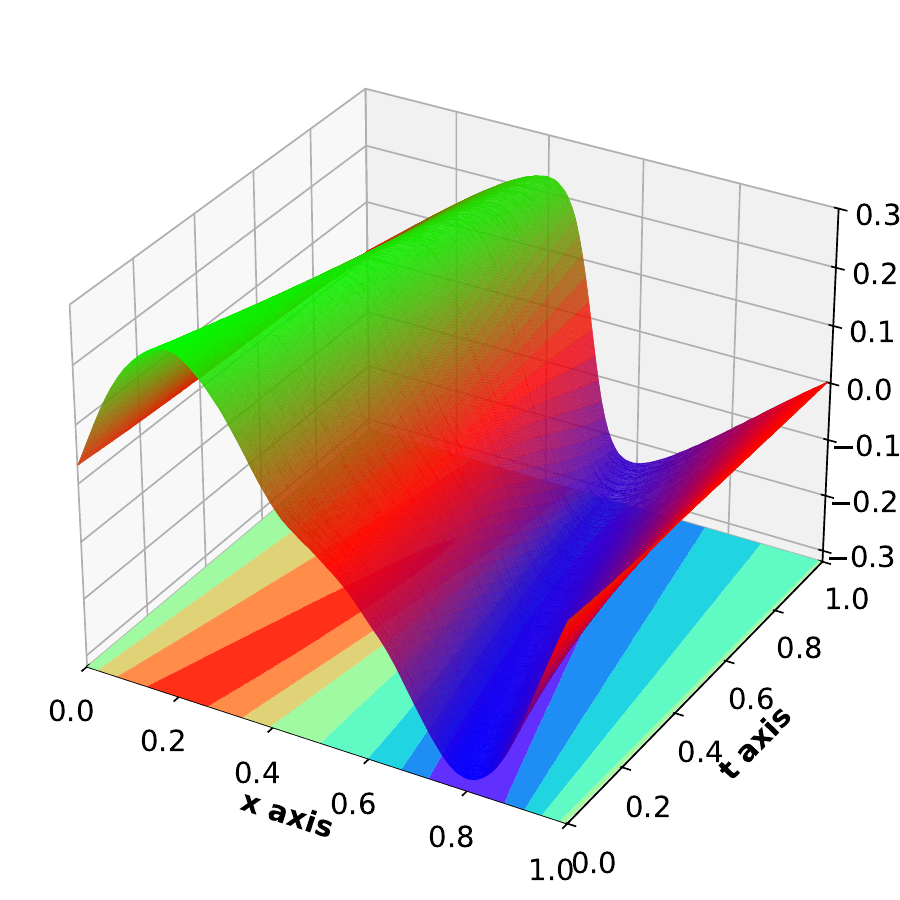} 
        \subcaption{MAD-RSNGS}
        \label{Fig.subfig_8_4}
    \end{minipage}
    
    \caption{Burgers equation: (a) analytic solution; (b) result of the MAD-PINN, which requires the full set of sampling points; (c) approximation obtained through MAD-NGM; (d) result of MAD-RSNGS, where only a subset of parameters are randomly updated. From the comparison of projections on the $x-t$ plane in Plots (a) and (b), discrepancies in projection patterns between the predictions of MAD-PINN and the true values are observed in the intervals $\ x \in [0.1,0.3]$ and $\ x \in [0.7,0.9]$, implying that the model still exhibits deficiencies in capturing dynamic changes within regions of strong gradients.} 
    \label{Fig.kdv_1str_burtgers_tttt}
\label{Fig.burgers_competition1}
\end{figure}

Figure~\ref{Fig.burgers_competition1} compares the numerical solutions obtained by the three methods in representative test samples. The three-dimensional plots, along with their projections onto the $x - t$ plane, illustrate the discrepancies between the predicted and reference solutions. It is evident that our methods produce solutions that are in closer agreement with the reference solutions, demonstrating improved accuracy and stronger generalization capability on unseen test data when compared with the MAD-PINN.

\begin{figure}[h] 
\centering 
\includegraphics[width=0.7\textwidth]{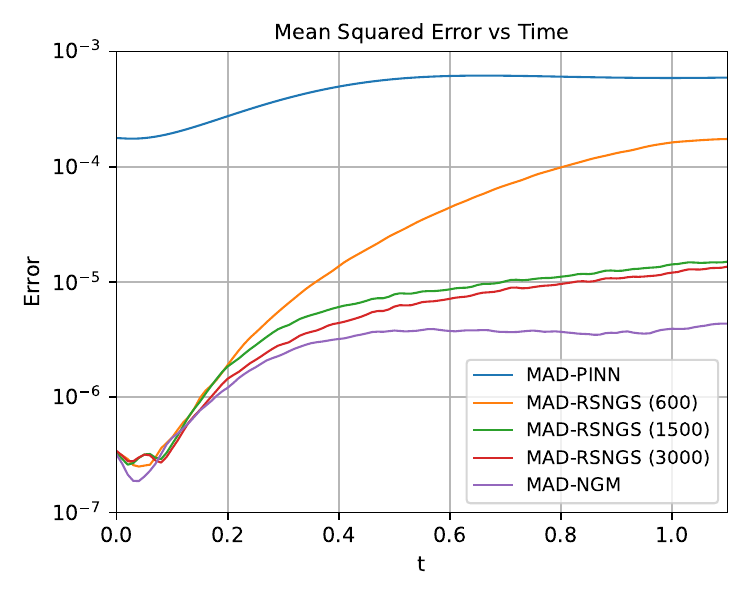} 
\caption{MSE of the approximation over the time interval $[0, 1.1]$, comparing the MAD-NGM, MAD-PINN and MAD-RSNGS. The MAD-PINN neglects temporal causality in the prediction process, resulting in severe error accumulation (around $10^{-4}$). In contrast, our method (particularly MAD-NGM) can correctly predict the results at all sampling points over a long-time domain, with the error reduced by 2 orders of magnitude (around $10^{-6}$).
} 
\label{Fig.burgers_competition} 
\end{figure}

\begin{table}[h]
    \centering
    \caption{Accuracy comparison of different methods at various time instants}
    \label{tab:table_three_methods_comparsion}
    \begin{tabular}{cccccc} 
    \toprule
    \multicolumn{1}{c}{} & $s$ & $t = 0.2$ & $t = 0.5$ & $t = 1.0$ & $t = 1.1$\\
    \midrule
    MAD-NGM &3811 & $1.18\times 10^{-6}$ & $3.91\times 10^{-6}$ & $4.11\times10^{-6}$ & $4.33\times 10^{-6}$\\
     \midrule
     \multirow{4}*{MAD-RSNGS} 
     & 600 &$1.92\times 10^{-6}$  & $2.62\times 10^{-5}$ & $1.62\times 10^{-4}$ & $1.74\times 10^{-4}$\\ 
     ~ & 1000 &$1.93\times 10^{-6}$  & $2.72\times 10^{-5}$ & $1.61\times 10^{-4}$ & $1.70\times 10^{-4}$\\ 
     ~ & 1500 &$1.91\times 10^{-6}$  & $8.11\times 10^{-6}$ & $1.45\times 10^{-5}$ &$2.54\times 10^{-5}$ \\
     ~ & 3000 &$1.51\times 10^{-6}$  & $6.34\times 10^{-6}$ & $1.21\times 10^{-5}$ &$1.41\times 10^{-5}$ \\ 
     \midrule
     MAD-PINN & $-$ & $2.75\times 10^{-4}$ & $5.76\times 10^{-4}$ & $5.89\times10^{-4}$ & $5.94\times10^{-4}$\\
    \bottomrule
    \end{tabular}%
\end{table}%

\begin{table}[h]
    \centering
    \caption{Average time cost (in minutes) of each method in the pre-training, fine-tuning, and time-evolution phases.}
    \label{tab:burger_time_comparison}
    \begin{tabular}{ccccc} 
    \toprule
    \multicolumn{1}{c}{} & $s$ & Pre-trained & Fine-tuned & Time-evolution \\
    \midrule
    MAD-NGM & 3811 & 44.16 & $7.21$ & $46.47$  \\
    \midrule
     \multirow{4}*{MAD-RSNGS} 
     & 600 & \multirow{4}*{44.16} & \multirow{4}*{$7.21$} & $3.50$  \\
     ~ & 1000 &  &  & $5.01$  \\
     ~ & 1500 &  &  & $5.24$  \\
     ~ & 3000 &  &  & $5.76$  \\
     \midrule
    MAD-PINN & $-$& $141.32$ & $4.03$ & $-$ \\
    \bottomrule
    \end{tabular}%
\end{table}

Based on the results presented in Figure~\ref{Fig.burgers_competition}, Table~\ref{tab:table_three_methods_comparsion} and Table~\ref{tab:burger_time_comparison}, the following observations can be made:

(1) As the number of randomly selected parameters $s$ per time step increases, the MAD-RSNGS exhibits a notable improvement in solution accuracy; however, its final accuracy remains approximately one order of magnitude lower than that achieved by the MAD-NGM method. In the context of parameterized PDEs, MAD-NGM consistently exhibits greater robustness and higher predictive accuracy than both MAD-RSNGS and MAD-PINN.

(2) A comparison of computational costs in Table~\ref{tab:burger_time_comparison} further reveals that MAD-RSNGS substantially reduces the time-evolution cost compared with MAD-NGM, while maintaining a comparable level of accuracy. MAD-PINN requires dense sampling over the entire spatiotemporal domain, which leads to large GPU memory consumption and a prolonged pre-training stage. Although its fine-tuning stage is relatively short, this efficiency is often achieved at the expense of accuracy, particularly for high-dimensional or strongly nonlinear parametric problems. In solving this problem, despite extensive optimization iterations and careful tuning of other hyperparameters, the accuracy achieved by MAD-PINN remains lower than that of the proposed methods.

The differences among the experimental results mainly come from the different methods used. Specifically:

(1) The PINN approach solves differential equations by embedding the governing physical constraints directly into the loss function, with the residuals evaluated at sampling points distributed throughout the entire space–time domain. However, this end-to-end training framework tends to accumulate systematic errors along the temporal direction, resulting in a noticeable degradation of accuracy for long-term predictions.

(2) The MAD-PINN inherits the aforementioned challenge of approximating the solution over the entire domain, while additionally requiring the model to learn diverse solution behaviors across a wide range of initial conditions. This substantially increases the burden on the model’s representational capacity. Consequently, improving its performance often necessitates either higher-dimensional latent vectors or finely tuned network parameters, both of which lead to greater computational complexity.

(3) In contrast, the proposed methods decouple the approximation of the initial condition from the time discretization, which allows efficient prediction beyond the training time range. With the same latent dimension and network structure, the MAD-RSNGS updates only a randomly selected subset of parameters at each time step, which significantly reduces the computational cost. However, since the MAD-NGM method updates all parameters at every step, it requires more computation time than MAD-RSNGS but achieves smaller errors at each time point, resulting in higher overall accuracy than the other two methods.

\subsection{Allen-Cahn  Equations} 
\label{allen-equation}
In this section, we investigate the generalization capability of the proposed method for solving parameterized partial differential equations, using the Allen–Cahn equation \cite{feng2003numerical} as a representative example. This nonlinear PDE models phase separation phenomena and is widely used in materials science to describe the evolution of phase boundaries.

The section is organized into two subsections, each addressing a different class of parameterized problems. The first subsection considers a one-dimensional Allen–Cahn equation, where both the initial condition and the spatial domain are influenced by random parameters. Numerical experiments are conducted under this setting to assess performance. The second subsection focuses on the solution of the two-dimensional Allen–Cahn equation, with an emphasis on analyzing the model's behavior and accuracy in higher-dimensional scenarios.

\subsubsection{One-dimensional Allen-Cahn Equation}

This subsection focuses on the case where both the solution domain and the initial condition are dependent on random parameters. Two sets of experiments are designed based on the same initial condition: one with time-independent parameters and another with time- and space-dependent parameters. These experiments aim to test the validity of the time-stepping approach.

Consider the example defined on the domain $\Omega = [0-\delta, 1+\delta], \delta \in [-0.2, 0.2]$ and time interval $T = [0, 2]$:

\begin{equation}
  \frac{\partial u}{\partial t} - 0.001 \frac{\partial^2 u}{\partial x^2} + a(t,x)(u^3 - u) = 0,
  \label{eq:ac_prev}
\end{equation}
which is subject to the following boundary and initial conditions:

\begin{itemize}
    \item[\textbf{(BC)}] Periodic boundary conditions: $u(-\delta,t) = u(1+\delta,t), \ t \in [0,2], \delta \in [-0.2, 0.2]$;
    \item[\textbf{(IC)}] Initial condition: $u(x,0) \sim N(0;{7^2}{(-\Delta + 7^2 I)^{-3}})$, where this random field satisfies the periodic boundary conditions.
\end{itemize}

To ensure that the approximate solution naturally satisfies the boundary conditions, the model input is chosen as:
$$
[\sin(\pi \frac{x+\delta}{1+2\delta}), \cos(\pi \frac{x+\delta}{1+2\delta}), \boldsymbol{z}].
$$
In this experiment, the latent vector dimension is set to 30, and a two-layer neural network with 30 neurons per layer is used for approximation. In the pre-training phase, the model is trained on 80 samples, and the L-BFGS optimizer is employed for 50,000 iterations. During the time update process, a sparse update scheme is adopted, where 1500 network parameters are randomly selected for each calculation. The reference solution is computed using a Fourier spectral discretization in space together with a fourth-order explicit Runge–Kutta scheme for time integration.

\paragraph{Time-independent Parameters}

Consider a more general form of the equation, where the parameter $a(t,x) = 2$. Then the equation \eqref{eq:ac_prev} becomes:
\begin{equation*}
\frac{\partial u}{\partial t} - 0.001 \frac{\partial^2 u}{\partial x^2} + 2(u^3 - u) = 0.
\end{equation*}

\begin{figure}[H] 
\centering 
    \begin{minipage}[c]{0.49\textwidth}
    \centering
    \includegraphics[width=\textwidth]{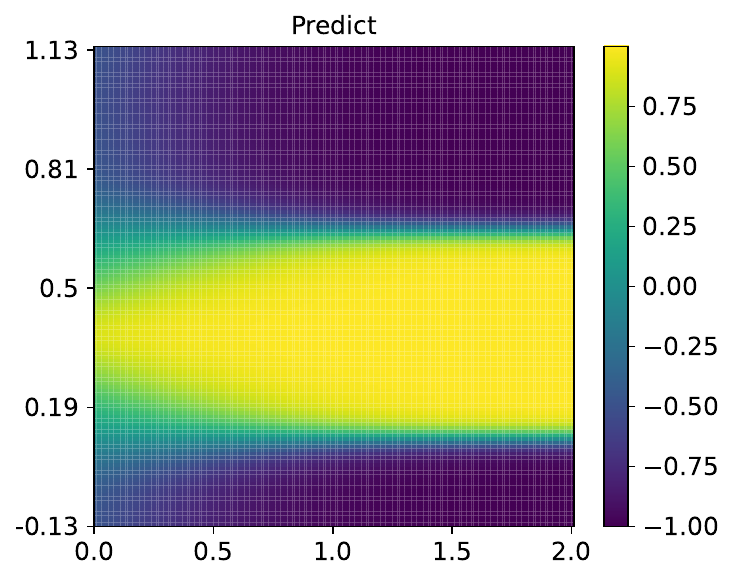}
    \label{Fig.subfig_10_1}
    \end{minipage}
    \centering
    \begin{minipage}[c]{0.49\textwidth}
    \centering
    \includegraphics[width=\textwidth]{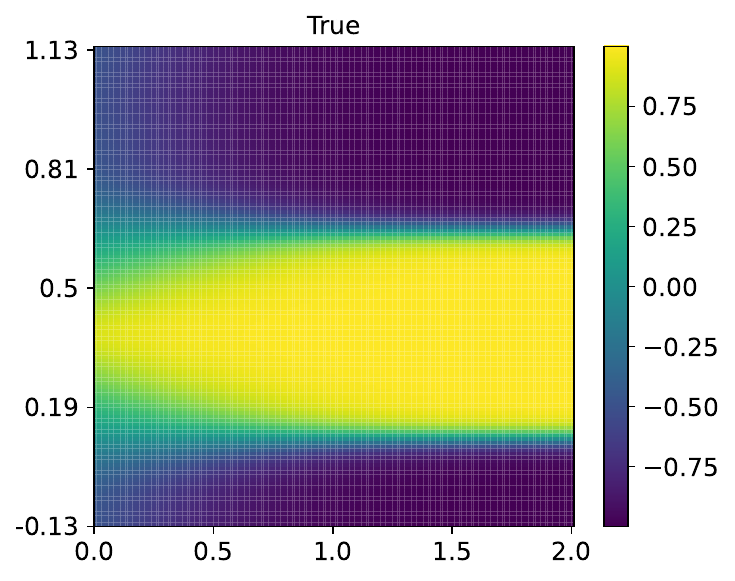}
    \label{subfig_10_2}
    \end{minipage}
\caption{Comparison of predicted and reference solutions for the one-dimensional AC equation. The striking visual concordance between the two panels highlights the method’s efficacy in faithfully reproducing the solution structure of the AC equation.} 
\label{Fig.Ac_TX83withOUTT} 
\end{figure}

\begin{figure}[H]
\centering 
    \begin{minipage}[c]{0.32\textwidth}
    \centering
    \includegraphics[width=\textwidth]{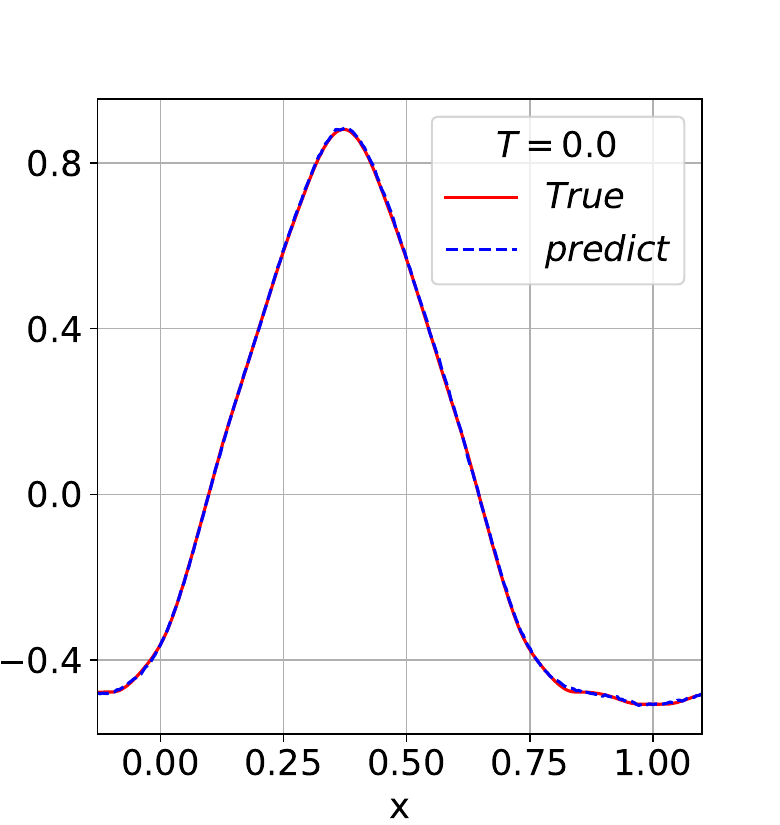}
    \end{minipage}
    \centering
    \begin{minipage}[c]{0.32\textwidth}
    \centering
    \includegraphics[width=\textwidth]{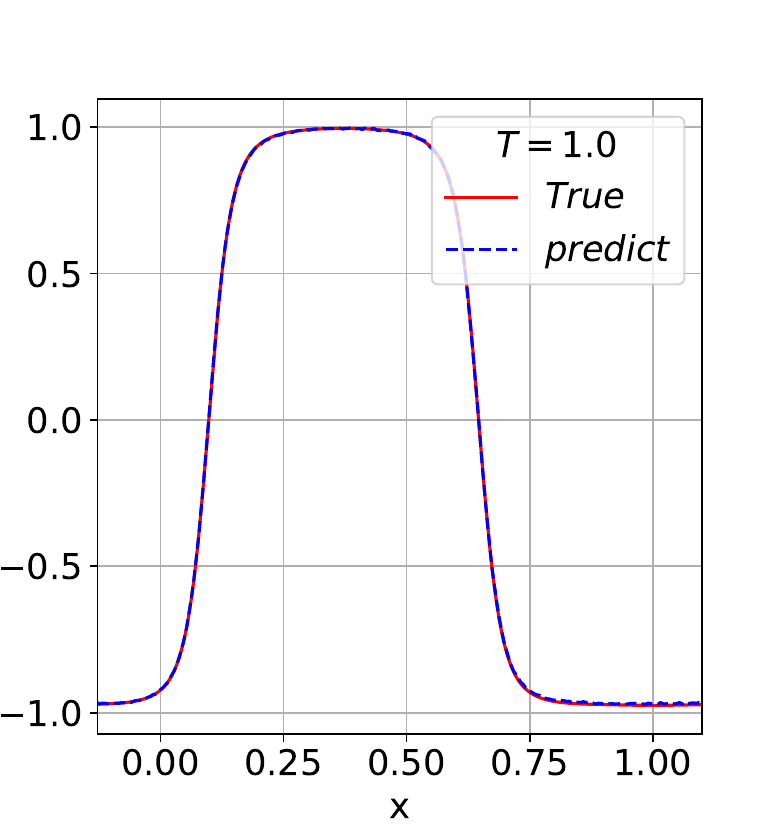}
    \end{minipage}
    \centering
    \begin{minipage}[c]{0.32\textwidth}
    \centering
    \includegraphics[width=\textwidth]{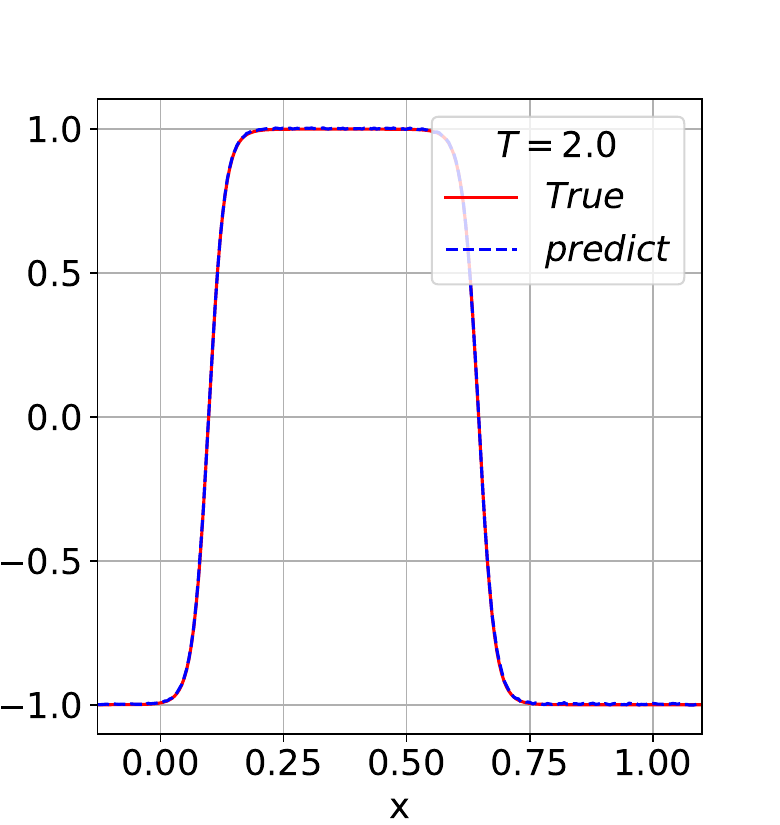}
    \end{minipage}\\
\caption{Temporal evolution of the reference and predicted solutions for the AC equation on the testing sample at $t = 0.0~,1.0~,2.0$. The predicted solutions accurately reproduce the characteristic phase separation and interface formation dynamics of the AC equation. } 
\label{Fig3.ac_withoutT_res_tttt} 
\end{figure}

A new sample from the test set was selected for fine-tuning, with 6,630 iterations performed using a time step of 0.001 until the terminal time $t=2$. The computational results, presented in Figure~\ref{Fig.Ac_TX83withOUTT}~\ref{Fig3.ac_withoutT_res_tttt}, indicate that for this new sample—whose solution domain approximately spans $[-0.128, 1.128]$.

\paragraph{Time-dependent Parameters}

\begin{figure}[H] 
\centering 
    \begin{minipage}[c]{0.49\textwidth}
    \centering
    \includegraphics[width=\textwidth]{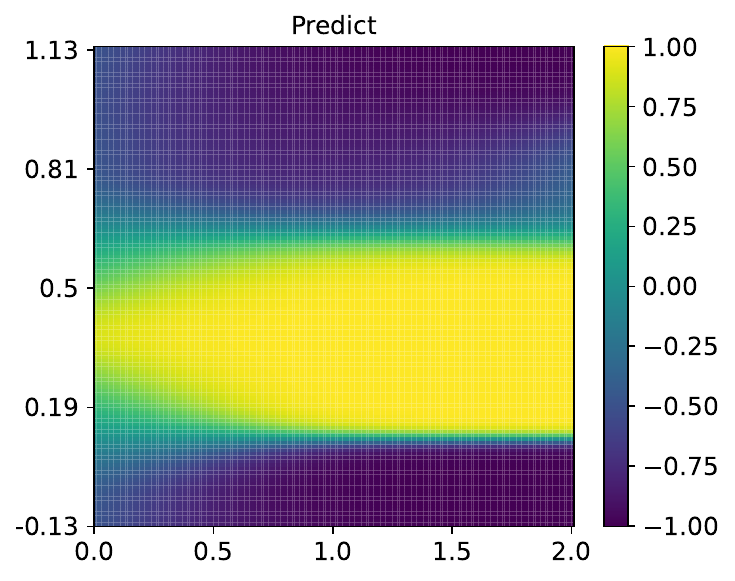}
    \label{Fig.subfig_12_1}
    \end{minipage}
    \centering
    \begin{minipage}[c]{0.49\textwidth}
    \centering
    \includegraphics[width=\textwidth]{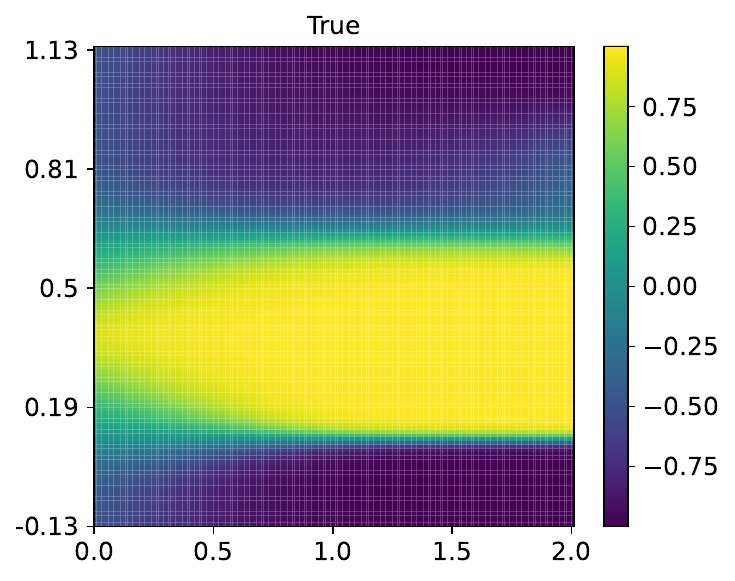}
    \label{subfig_12_2}
    \end{minipage}
\caption{Comparison of predicted and reference solutions for the one-dimensional AC equation in the time-dependent parameter case. Despite time-dependent parameters causing dynamic changes in the phase-separation process, the predicted solution highlights the method's effectiveness in capturing the transient solution structure of the AC equation.} 
\label{Fig.Ac_TX83withT} 
\end{figure}

\begin{figure}[H]
\centering 
    \begin{minipage}[c]{0.32\textwidth}
    \centering
    \includegraphics[width=\textwidth]{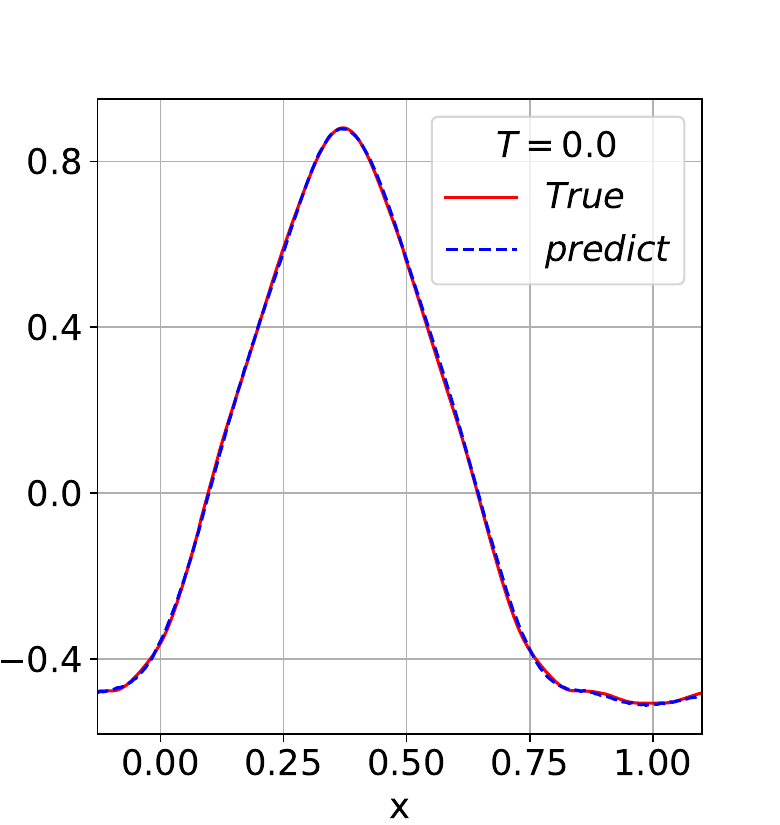}
    \end{minipage}
    \centering
    \begin{minipage}[c]{0.32\textwidth}
    \centering
    \includegraphics[width=\textwidth]{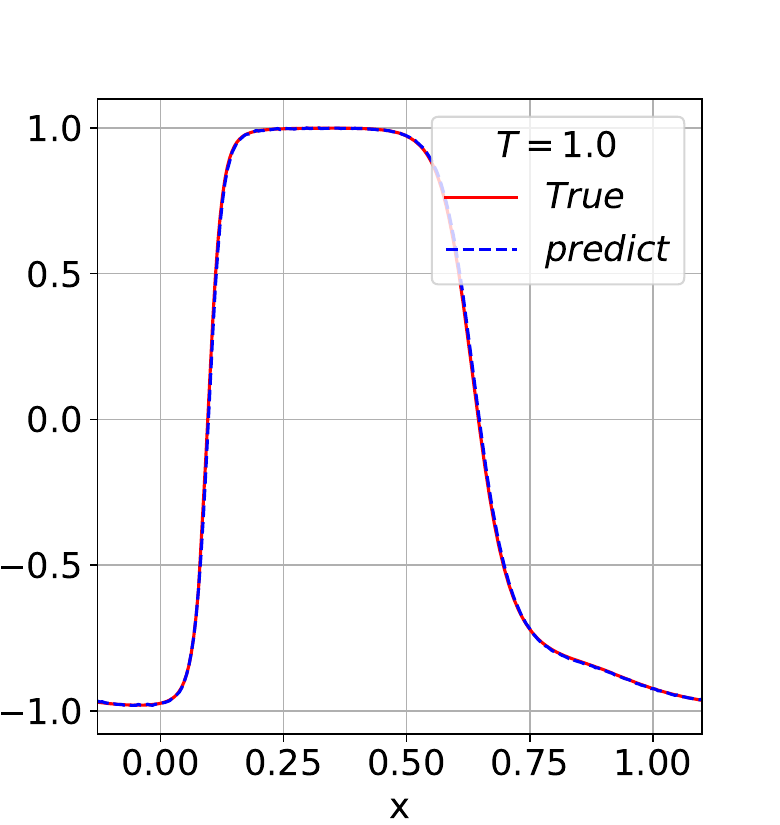}
    \end{minipage}
    \centering
    \begin{minipage}[c]{0.32\textwidth}
    \centering
    \includegraphics[width=\textwidth]{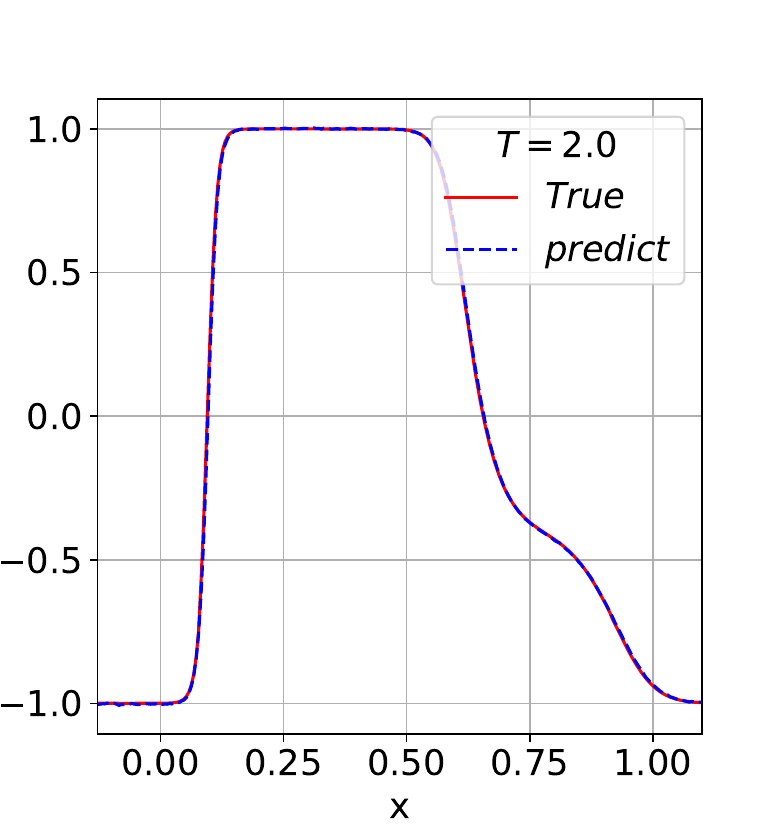}
    \end{minipage}\\
\caption{Temporal evolution of the reference and predicted solutions for the AC equation with Time-Varying Parameters at $t = 0.0~,1.0~,2.0$.} 
\label{Fig3.ac_1d_withT_res_tttt} 
\end{figure}

Consider the general form of the equation \eqref{eq:ac_prev}, where the parameter term $a(x,t)$ is related to both time and space:
\begin{equation*}
\frac{\partial u}{\partial t}-0.001\frac{\partial^2u}{\partial x^2}+2[1+t\sin(2 \pi \frac{x+\delta}{1+2\delta})]\cdot(u^3 - u) = 0 .
\end{equation*}
The same model parameters and initial conditions are employed in this setting. The calculation results are shown in Figure\ref{Fig.Ac_TX83withT} and \ref{Fig3.ac_1d_withT_res_tttt}, which approximately span $[-0.128, 1.128]$. This change makes the parameter term $a(x,t)$ vary with time $t$ and spatial position $x$, thus introducing a more complex spatiotemporal dependence. 

In order to further analyze the evolution process and the behavior of the solution of the equation in different time steps, compare the approximate errors of the solution at different time points in the two cases of parameter terms, and the detailed calculation results are shown in Table~\ref{tab:ac_tx}: 

\begin{table}[h]
    \centering
    \caption{Approximate errors of the MAD-RSNGS in the two cases of parameter terms at $ t = 0.4,\, 1.0,\, 2.0 $.}
    \label{tab:ac_tx}
    \begin{tabular}{ccccc}
    \toprule
      \multicolumn{1}{c}{Parameter term setting} &$t = 0.4$&$t = 1.0$&$t = 2.0$\\
         \midrule
      Time-independent  & $2.68\times 10^{-5}$ &$5.18\times10^{-5}$  &$7.35\times 10^{-5}$  \\
     \midrule
     Time-dependent& $5.35\times 10^{-5}$ &$6.98\times10^{-5}$  &$6.78\times 10^{-5}$  \\
 \bottomrule
    \end{tabular}%
\end{table}%

The experimental results show that: (1) By constructing the network input structure of the meta-learning self-decoder, the model can also be applied to relevant problems in the random solution region; (2) For this example, regardless of whether the parameter terms in the equation are related to time and space, the adopted method can achieve a good approximation of the reference solution. 

\subsubsection{Two-dimensional Allen-Cahn Equation}
Consider the equation with the solution region $\Omega = [0,1]^2$ and time interval $T = [0,2]$ in the two-dimensional case:
\begin{equation*}
\frac{\partial u}{\partial t}-0.001\Delta  u+2(u^3 - u) = 0  ,
\end{equation*}
which satisfies the following boundary and initial conditions:
\begin{itemize}
    \item[\textbf{(BC)}] Periodic boundary conditions: $u(0,y,t) = u(1,y,t),u(x,0,t)=u(x,1,t),t \in [0,2]$;
    \item[\textbf{(IC)}] Initial condition: The initial value condition is considered as a combination of trigonometric functions with random coefficients, such as:
\end{itemize} 
\begin{equation*}
u(x,y,0) = 0.001\sum_{i=-1}^{1}\sum_{j=-1}^{1}[\alpha_{i,j}\sin(2\pi i x+2\pi i y)+\beta_{i,j}\cos(2\pi i x+2\pi j y)].
\label{equ:double_ac}
\end{equation*}
Here, by introducing random coefficients $\alpha_{i,j}\in[0,1]$ and $\beta_{i,j} \in[0,1]$, we can simulate the evolution of the system under different initial states.
For example, when studying problems such as phase transition processes in materials and morphogenesis in biological tissues, the initial state of the system may have a spatial distribution characteristic similar to that described by the combination of trigonometric functions, or there may be certain random perturbations.

\begin{figure}[H] 
\centering 
    \begin{minipage}[t]{0.32\textwidth}
    \centering
    \includegraphics[width=\textwidth]{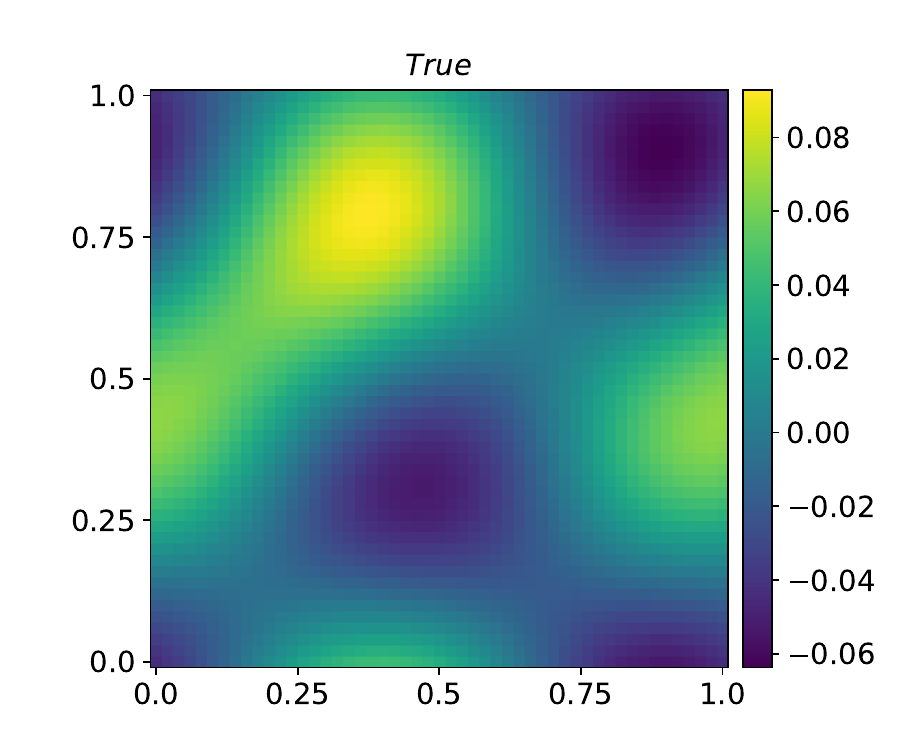}
    \label{Fig.subfig_14_1}
    \end{minipage}
    \begin{minipage}[t]{0.32\textwidth}
    \centering
    \includegraphics[width=\textwidth]{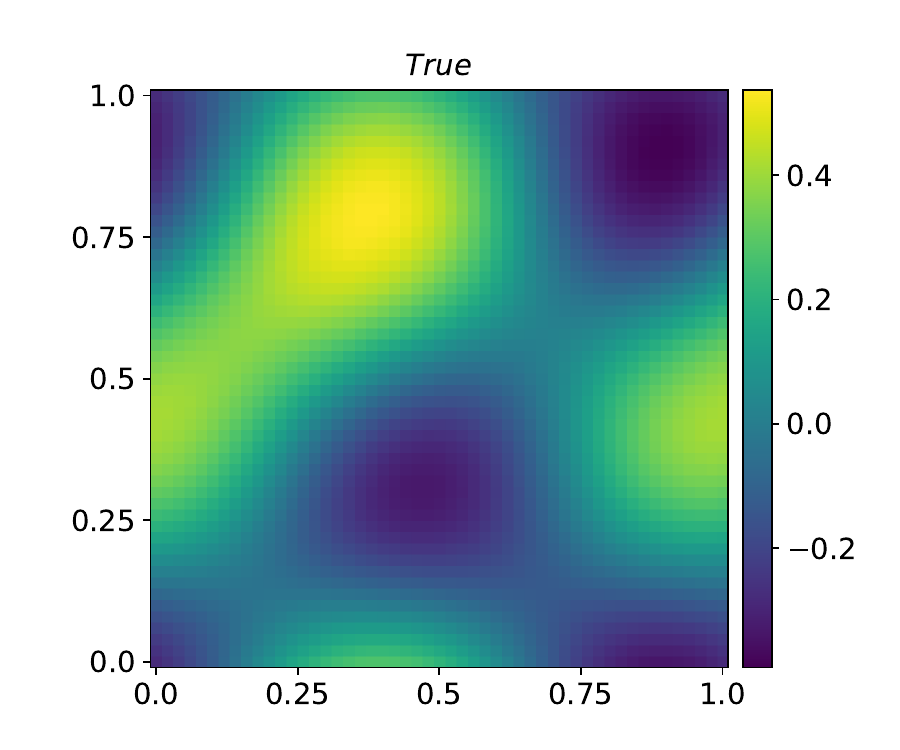}
    \label{subfig_14_2}
    \end{minipage}
    \begin{minipage}[t]{0.32\textwidth}
    \centering
    \includegraphics[width=\textwidth]{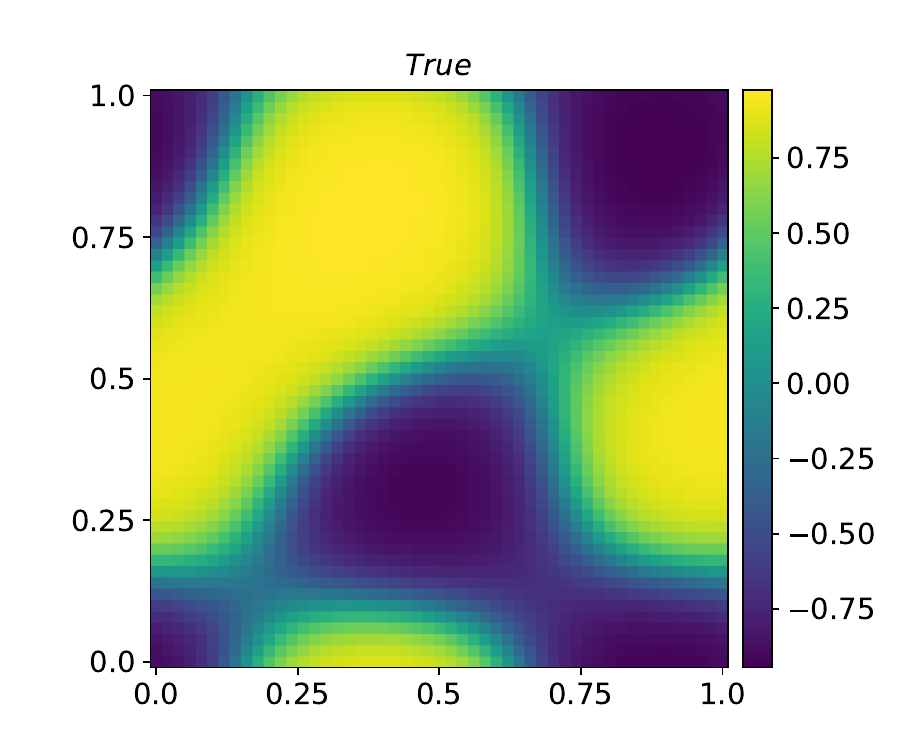}
    \label{Fig.subfig_14_3}
    \end{minipage}
    \vspace{0.5cm}
    \begin{minipage}[t]{0.32\textwidth}
    \centering
    \includegraphics[width=\textwidth]{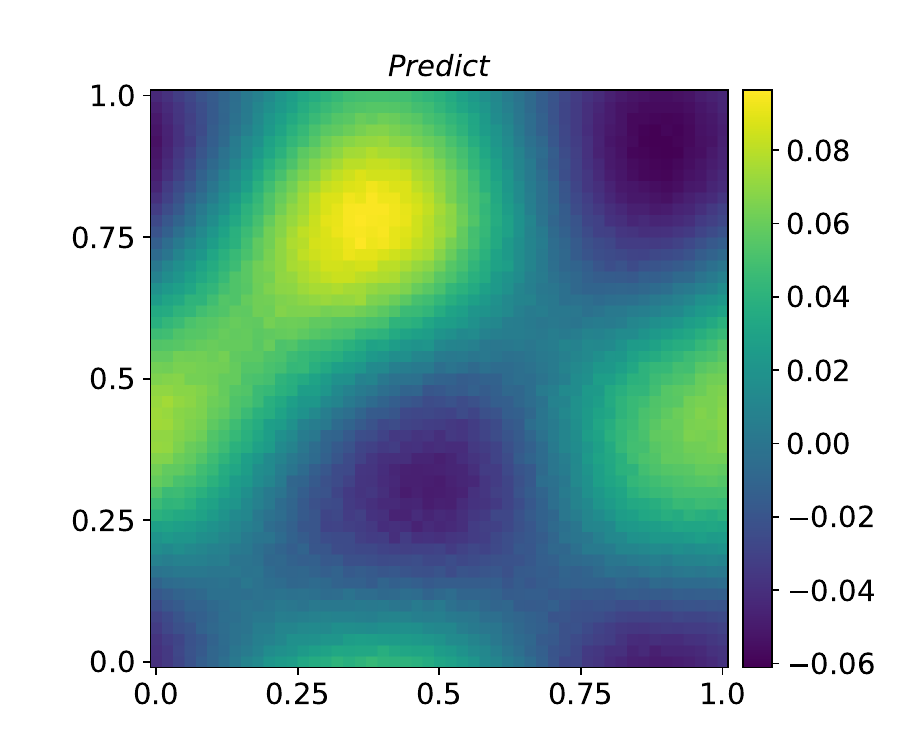}
    \subcaption{$t = 0.0$}
    \label{subfig_14_4}
    \end{minipage}
    \begin{minipage}[t]{0.32\textwidth}
    \centering
    \includegraphics[width=\textwidth]{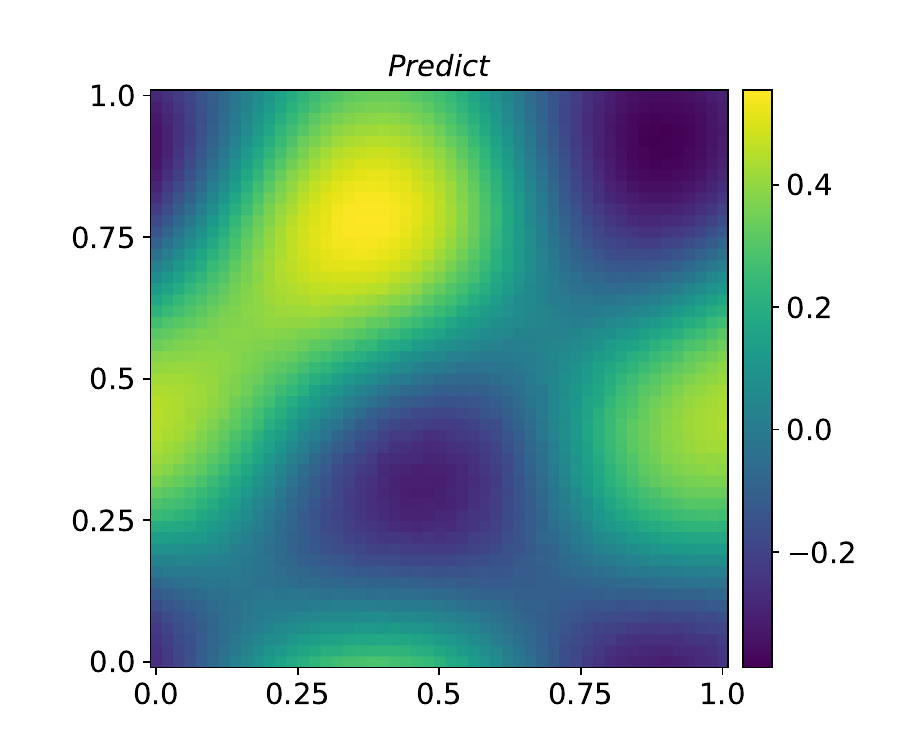}
    \subcaption{$t = 1.0$}
    \label{Fig.subfig_14_5}
    \end{minipage}
    \begin{minipage}[t]{0.32\textwidth}
    \centering
    \includegraphics[width=\textwidth]{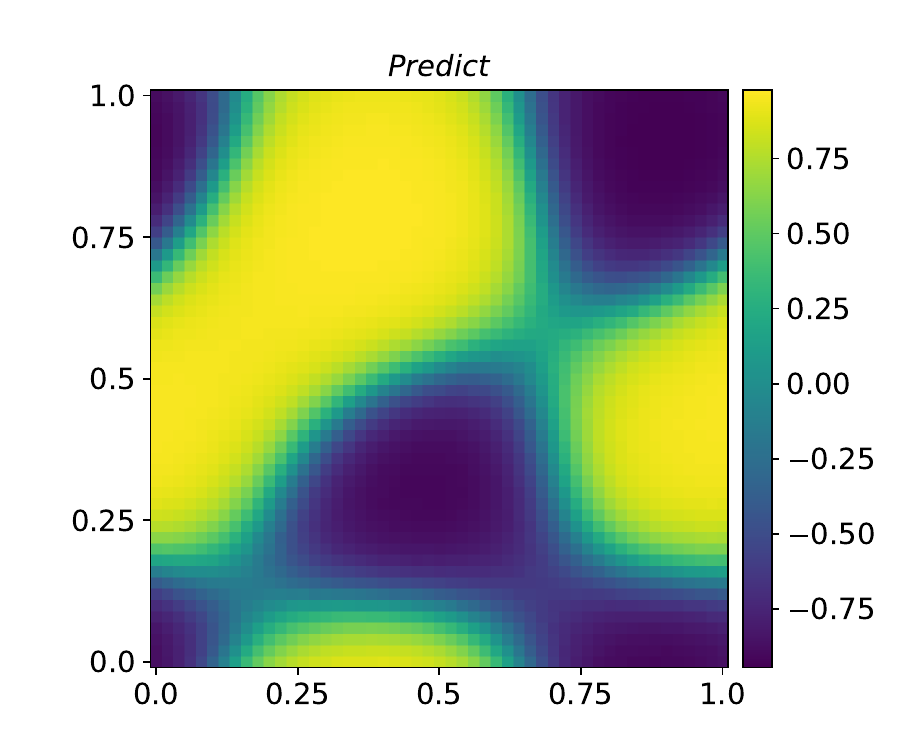}
    \subcaption{$t = 2.0$}
    \label{subfig_14_6}
    \end{minipage}
\caption{Approximate and reference solutions of the two-dimensional Allen-Cahn equation at the time points $t = 0.0,\ 1.0,\ 2.0$. This result substantiates that our method retains robust efficacy in solving high-dimensional partial differential PDE problems, transcending the constraints of low-dimensional scenarios.} 
\label{Fig.ac2d0_res_ng} 
\end{figure}

In this experiment, the dimension of the hidden vector is 60, and an eight-layer neural network is used, with 30 neurons in each layer. Pre-training is carried out based on 180 samples. The L-BFGS optimizer is used for 100,000 iterations. In the fine-tuning optimization stage, the Adam optimizer is used for 8,000 training iterations. In the time update process of the sparse scheme, the time step is set to 0.001, and the iteration is updated until the termination time $t = 2.0$. At each time step, 1500 parameters are randomly selected to participate in the calculation.
The approximate results at the time points $t = 0.0,~1.0,~2.0$ are taken and plotted in Figure~\ref{Fig.ac2d0_res_ng}.

\begin{figure}[H] 
\centering 
\includegraphics[width=0.7\textwidth]{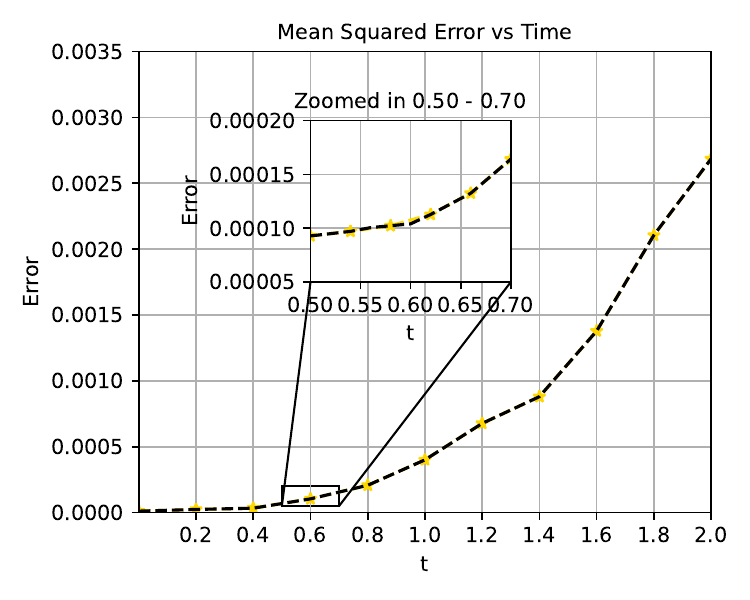} 
\caption{MSE of the approximate solution over the time interval $[0, 2]$. Notably, at $t=2.0$ , the prediction error remains as low as 0.0026. This result demonstrates the MAD-RSNGS’s efficacy in sustaining accuracy over extended time horizons, thereby evidencing its potential for addressing high-dimensional, long-time dynamical problems. } 
\label{Fig.ac2d_0_res_ng} 
\end{figure}
As shown in Figure \ref{Fig.ac2d_0_res_ng}, the computational complexity tends to increase significantly when addressing higher-dimensional problems. In this complex environment, the method proposed in this paper can accurately and efficiently capture the key features of the solution of the equation by virtue of its internal mechanism. Whether it is the slight fluctuations of the solution or its overall distribution trend, it can be keenly observed by this method. This shows that the MAD-RSNGS has good adaptability and can more accurately capture the characteristics of the solution of the equation in the high-dimensional space. This result further proves the potential of the MAD-RSNGS for high-dimensional problems.

\section{Conclusions}
\label{sec:conclusions}
This paper addresses the challenge of efficiently solving parametric, time-dependent PDEs, where a major difficulty lies in managing the uncertainty arising from varying initial conditions, often resulting in high computational costs. To overcome this, we propose an enhanced Neural Galerkin framework that incorporates the MAD paradigm. This framework can model the encountered uncertainty and quickly generate appropriate initial neural network parameters, then leverage space-time decoupling and introduce a randomized sparse update strategy to achieve accurate and efficient approximation of PDE solutions.

We validated the effectiveness and strong generalization ability of the proposed method through several numerical examples, including the Korteweg–de Vries equation, the Burgers equation, and the one- and two-dimensional Allen–Cahn equations. In all cases, under the same network architecture and parameter settings, the MAD-NGM algorithm and its extended version MAD-RSNGS, produced approximate solutions that successfully preserved the physical properties of the partial differential equations.

With the continuous growth of computational resources and the advancement of deep learning techniques, we expect to extend this approach to high-dimensional and complex problems, particularly in real-time simulations and large-scale data processing, to further verify its generalization performance in more challenging scenarios. In addition, considering the current sensitivity of the method to network architecture and hyperparameter choices, future work will focus on developing adaptive network designs and multi-level optimization strategies to further improve the robustness and practical applicability of the method.

\section*{Acknowledgments}

The research of this work was supported by the National Natural Science Foundation of China (No. 12301520, 12471405), JR25003, the Beijing Outstanding Young Scientist Program (No. JWZQ20240101027), and the Science and Technology Innovation Program of Hunan Province (No. 2025RC3080). 

\bibliographystyle{elsarticle-num} 
\bibliography{main}

\begin{thebibliography}{10}
\expandafter\ifx\csname url\endcsname\relax
  \def\url#1{\texttt{#1}}\fi
\expandafter\ifx\csname urlprefix\endcsname\relax\def\urlprefix{URL }\fi
\expandafter\ifx\csname href\endcsname\relax
  \def\href#1#2{#2} \def\path#1{#1}\fi

\bibitem{cybenko1989approximation}
G.~Cybenko, Approximation by superpositions of a sigmoidal function, Mathematics of Control, Signals and Systems 2 (1989) 303--314.

\bibitem{hornik1989multilayer}
K.~Hornik, M.~Stinchcombe, H.~White, Multilayer feedforward networks are universal approximators, Neural Networks 2~(5) (1989) 359--366.

\bibitem{raissi2019physics}
M.~Raissi, P.~Perdikaris, G.~E. Karniadakis, Physics-informed neural networks: A deep learning framework for solving forward and inverse problems involving nonlinear partial differential equations, Journal of Computational Physics 378 (2019) 686--707.

\bibitem{sirignano2018dgm}
J.~Sirignano, K.~Spiliopoulos, Dgm: A deep learning algorithm for solving partial differential equations, Journal of Computational Physics 375 (2018) 1339--1364.

\bibitem{yu2018deep}
W.~E, B.~Yu, The deep ritz method: A deep learning-based numerical algorithm for solving variational problems, Communications in Mathematics and Statistics 6~(1) (2018) 1--12.

\bibitem{shan2025vi}
B.~Shan, Y.~Li, S.~Huang, Vi-pinns: Variance-involved physics-informed neural networks for fast and accurate prediction of partial differential equations, Neurocomputing 623 (2025) 129360.

\bibitem{zhang2020learning}
D.~Zhang, L.~Guo, G.~E. Karniadakis, Learning in modal space: Solving time-dependent stochastic pdes using physics-informed neural networks, SIAM Journal on Scientific Computing 42~(2) (2020) A639--A665.

\bibitem{han2018solving}
J.~Han, A.~Jentzen, W.~E, Solving high-dimensional partial differential equations using deep learning, Proceedings of the National Academy of Sciences 115~(34) (2018) 8505--8510.

\bibitem{zhang2018deep}
L.~Zhang, J.~Han, H.~Wang, R.~Car, W.~E, Deep potential molecular dynamics: a scalable model with the accuracy of quantum mechanics, Physical review letters 120 (2018) 143001.

\bibitem{weinan2021algorithms}
W.~Ee, J.~Han, A.~Jentzen, Algorithms for solving high dimensional pdes: from nonlinear monte carlo to machine learning, Nonlinearity 35~(1) (2021) 278--310.

\bibitem{zang2020weak}
Y.~Zang, G.~Bao, X.~Ye, H.~Zhou, Weak adversarial networks for high-dimensional partial differential equations, Journal of Computational Physics 411 (2020) 109409.

\bibitem{dissanayake1994neural}
M.~G. Dissanayake, N.~Phan-Thien, Neural-network-based approximations for solving partial differential equations, Communications in Numerical Methods in Engineering 10~(3) (1994) 195--201.

\bibitem{krishnapriyan2021characterizing}
A.~S. Krishnapriyan, A.~Gholami, S.~Zhe, R.~M. Kirby, M.~W. Mahoney, Characterizing possible failure modes in physics-informed neural networks, Advances in neural information processing systems 34 (2021) 26548--26560.

\bibitem{dirac1930note}
P.~A. Dirac, Note on exchange phenomena in the thomas atom, Mathematical proceedings of the Cambridge philosophical society 26~(3) (1930) 376--385.

\bibitem{du2021evolutional}
Y.~Du, T.~A. Zaki, Evolutional deep neural network, Physical Review E 104 (2021) 045303.

\bibitem{bruna2024neural}
J.~Bruna, B.~Peherstorfer, E.~Vanden-Eijnden, Neural galerkin schemes with active learning for high-dimensional evolution equations, Journal of Computational Physics 496 (2024) 112588.

\bibitem{anderson2022evolution}
W.~Anderson, M.~Farazmand, Evolution of nonlinear reduced-order solutions for pdes with conserved quantities, SIAM Journal on Scientific Computing 44~(1) (2022) A176--A197.

\bibitem{berman2024neural}
J.~Berman, P.~Schwerdtner, B.~Peherstorfer, Neural galerkin schemes for sequential-in-time solving of partial differential equations with deep networks, in: Handbook of Numerical Analysis, Vol.~25, Elsevier, 2024, pp. 389--418.

\bibitem{berman2023randomized}
J.~Berman, B.~Peherstorfer, Randomized sparse neural galerkin schemes for solving evolution equations with deep networks, Advances in Neural Information Processing Systems (2023) 4097--4114.

\bibitem{kast2024positional}
M.~Kast, J.~S. Hesthaven, Positional embeddings for solving pdes with evolutional deep neural networks, Journal of Computational Physics 508 (2024) 112986.

\bibitem{ning2025filtered}
Z.~Ning, B.~Peherstorfer, Filtered neural galerkin model reduction schemes for efficient propagation of initial condition uncertainties in digital twins, arXiv preprint arXiv:2511.00670 (2025).

\bibitem{lu2019deeponet}
L.~Lu, P.~Jin, G.~Pang, Z.~Zhang, G.~E. Karniadakis, Learning nonlinear operators via deeponet based on the universal approximation theorem of operators, Nature Machine Intelligence 3~(3) (2021) 218--229.

\bibitem{li2020fourier}
Z.~Li, N.~Kovachki, K.~Azizzadenesheli, B.~Liu, K.~Bhattacharya, A.~Stuart, A.~Anandkumar, Fourier neural operator for parametric partial differential equations, arXiv preprint arXiv:2010.08895 (2020).

\bibitem{raonic2023convolutional}
B.~Raonic, R.~Molinaro, T.~Rohner, S.~Mishra, E.~de~Bezenac, Convolutional neural operators, in: ICLR 2023 workshop on physics for machine learning, 2023.

\bibitem{long2018pde}
Z.~Long, Y.~Lu, X.~Ma, B.~Dong, Pde-net: Learning pdes from data, in: International conference on machine learning, PMLR, 2018, pp. 3208--3216.

\bibitem{long2019pde}
Z.~Long, Y.~Lu, B.~Dong, Pde-net 2.0: Learning pdes from data with a numeric-symbolic hybrid deep network, Journal of Computational Physics 399 (2019) 108925.

\bibitem{jin2022mionet}
P.~Jin, S.~Meng, L.~Lu, Mionet: Learning multiple-input operators via tensor product, SIAM Journal on Scientific Computing 44~(6) (2022) A3490--A3514.

\bibitem{li2024physics}
Z.~Li, H.~Zheng, N.~Kovachki, D.~Jin, H.~Chen, B.~Liu, K.~Azizzadenesheli, A.~Anandkumar, Physics-informed neural operator for learning partial differential equations, ACM/IMS Journal of Data Science 1~(3) (2024) 1--27.

\bibitem{hao2023gnot}
Z.~Hao, Z.~Wang, H.~Su, C.~Ying, Y.~Dong, S.~Liu, Z.~Cheng, J.~Song, J.~Zhu, Gnot: A general neural operator transformer for operator learning, in: International Conference on Machine Learning, PMLR, 2023, pp. 12556--12569.

\bibitem{bhattacharya2021model}
K.~Bhattacharya, B.~Hosseini, N.~B. Kovachki, A.~M. Stuart, Model reduction and neural networks for parametric pdes, The SMAI journal of computational mathematics 7 (2021) 121--157.

\bibitem{Chen2024Positional}
J.~Chen, K.~Wu, Positional knowledge is all you need: Position-induced transformer (pit) for operator learning, arXiv preprint arXiv:2405.09285 (2024).

\bibitem{Huang202X}
J.~Huang, R.~You, T.~Zhou, Deep learning for the semi-classical limit of the schr$\backslash$" odinger equation, arXiv preprint arXiv:2509.04453 (2025).

\bibitem{Feng202X}
X.~Feng, L.~Guo, X.~Wan, H.~Wu, T.~Zhou, W.~Zhou, Lvm-gp: Uncertainty-aware pde solver via coupling latent variable model and gaussian process, arXiv preprint arXiv:2507.22493 (2025).

\bibitem{chen2025tensor}
Y.~Chen, Y.~Lin, X.~Sun, C.~Yuan, Z.~Gao, Tensor decomposition-based neural operator with dynamic mode decomposition for parameterized time-dependent problems, Journal of Computational Physics 533 (2025) 113996.

\bibitem{Venturi2023SVD}
S.~Venturi, T.~Casey, Svd perspectives for augmenting deeponet flexibility and interpretability, Computer Methods in Applied Mechanics and Engineering 403 (2023) 115718.

\bibitem{psaros2022meta}
A.~F. Psaros, K.~Kawaguchi, G.~E. Karniadakis, Meta-learning pinn loss functions, Journal of Computational Physics 458 (2022) 111121.

\bibitem{finn2017model}
C.~Finn, P.~Abbeel, S.~Levine, Model-agnostic meta-learning for fast adaptation of deep networks, in: International Conference On Machine Learning, Vol.~70, PMLR, 2017, pp. 1126--1135.

\bibitem{antoniou2018train}
A.~Antoniou, H.~Edwards, A.~Storkey, How to train your maml, arXiv preprint arXiv:1810.09502 (2018).

\bibitem{nichol2018reptile}
A.~Nichol, J.~Schulman, J.~Achiam, On first-order meta-learning algorithms, arXiv preprint arXiv:1803.02999 (2018).

\bibitem{yoon2018bayesian}
J.~Yoon, T.~Kim, O.~Dia, S.~Kim, Y.~Bengio, S.~Ahn, Bayesian model-agnostic meta-learning, Advances in neural information processing systems 31 (2018).

\bibitem{ye2024meta}
Z.~Ye, X.~Huang, H.~Liu, B.~Dong, Meta-auto-decoder: a meta-learning-based reduced order model for solving parametric partial differential equations, Communications on Applied Mathematics and Computation 6~(2) (2024) 1096--1130.

\bibitem{aproximate_pde}
A.~Cohen, R.~DeVore, Approximation of high-dimensional parametric pdes, Acta Numerica 24 (2015) 1--159.

\bibitem{devore1989optimal}
R.~A. DeVore, R.~Howard, C.~Micchelli, Optimal nonlinear approximation, Manuscripta mathematica 63 (1989) 469--478.

\bibitem{miles1981korteweg}
J.~W. Miles, The korteweg-de vries equation: a historical essay, Journal of Fluid Mechanics 106 (1981) 131--147.

\bibitem{butcher1996history}
J.~C. Butcher, A history of runge-kutta methods, Applied numerical mathematics 20~(3) (1996) 247--260.

\bibitem{bonkile2018systematic}
M.~P. Bonkile, A.~Awasthi, C.~Lakshmi, V.~Mukundan, V.~Aswin, A systematic literature review of burgers’ equation with recent advances, Pramana 90 (2018).

\bibitem{feng2003numerical}
X.~Feng, A.~Prohl, Numerical analysis of the allen-cahn equation and approximation for mean curvature flows, Numerische Mathematik 94 (2003) 33--65.

\end{thebibliography}

\end{document}